# Dynamic Gaussian Graph Operator: Learning parametric partial differential equations in arbitrary discrete mechanics problems

Chu Wang[1], Jinhong Wu[1], Yanzhi Wang[1], Zhijian Zha[1], Qi Zhou[1,*]

1. School of Aerospace Engineering, Huazhong University of Science & Technology, Wuhan, Hubei, 430074, China

**Abstract**

Deep learning methods have access to be employed for solving physical systems governed by parametric partial differential equations (PDEs) due to massive scientific data. It has been refined to operator learning that focuses on learning non-linear mapping between infinite-dimensional function spaces, offering interface from observations to solutions. However, state-of-the-art neural operators are limited to constant and uniform discretization, thereby leading to deficiency in generalization on arbitrary discretization schemes for computational domain. In this work, we propose a novel operator learning algorithm, referred to as Dynamic Gaussian Graph Operator (DGGO) that expands neural operators to learning parametric PDEs in arbitrary discrete mechanics problems. The Dynamic Gaussian Graph (DGG) kernel learns to map the observation vectors defined in general Euclidean space to metric vectors defined in high-dimensional uniform metric space. The DGG integral kernel is parameterized by Gaussian kernel weighted Riemann sum approximating and using dynamic message passing graph to depict the interrelation within the integral term. Fourier Neural Operator is selected to localize the metric vectors on spatial and frequency domains. Metric vectors are regarded as located on latent uniform domain, wherein spatial and spectral transformation offer highly regular constraints on solution space. The efficiency and robustness of DGGO are validated by applying it to solve numerical arbitrary discrete mechanics problems in comparison with mainstream neural operators. Ablation experiments are implemented to demonstrate the effectiveness of spatial transformation in the DGG kernel. The proposed method is utilized to forecast stress field of hyper-elastic material with geometrically variable void as engineering application.

**Keywords:** Scientific machine learning; Operator learning; Parametric partial differential equations; Gaussian kernel; Graph neural networks

*Corresponding author: qizhou@hust.edu.cn && qizhouhust@gmail.com (Qi Zhou)

# 1 Introduction

Partial differential equations (PDEs) are widely existing for governing physical systems. Significant in-depth research has been conducted to depict intricate dynamic problems, e.g., fluid flow, thermal advection, and solid mechanics. Due to the scarcity of acquiring analytical solutions for a series of PDEs, numerical techniques applied for solving PDEs have been proposed in the past decades such as finite elements method [1], finite difference method [2], and finite volume method [3] with their enhanced versions [4-6]. Nonetheless, two primary drawbacks are concomitant with the application of the aforementioned methods. First, formulating the exact parametric PDEs that contain multiple physical quantities necessitates massive prior knowledge which leads to plenty of complexity in numerical computation. The prior knowledge can be neither utterly acquired for all scenarios nor deduced mutually, especially for complicated multi-physics systems governed by complex parametric PDEs. For multi-physics systems like structure-acoustic coupling as an example [7], physical quantities of initial condition involve frequency, elasticity modulus and density. All quantities dynamically form the changeable parametric conditions for the system. Grasping all physical quantities pervasively for numerical methods is elusive. Second, numerically solving multi-physics systems governed by non-linear PDEs is highly time-consuming. The construction of spatiotemporal discrete model of computational domain is an extensive workload and numerical computation are necessarily implemented repetitively when dealing with new physical conditions for PDEs.

Deep neural networks can be employed to characterize mathematical equations derived from physical applications [8], hence, tackling the above two drawbacks. The ability of non-linear mapping between the input space and output space facilitates networks to realize rapidly generating solutions of parametric PDEs based on the given formulations. In recent years, study of neural network architecture has been emergent in processing the increasing bulk of data originating from numerical computation or experiments. Utilization of neural networks mainly operates in two mechanisms denoted as supervised learning and unsupervised learning. In the supervised regime, the learning of neural networks is driven by the prior knowledge and the ground true solutions combined as pair of data [9-13]. The input of neural networks is composed of separate computational conditions and quantities in the governing PDE of a physical field. The output is either the pointwise solutions of computational domain [14-16] or pixels of physics image [17-19]. In the unsupervised or weakly-supervised regime, referred to as physics-informed learning [20-24], the neural networks are trained with the constraint of PDE and a minority of prior solutions. In contrast to the data-driven regime, the values of PDE are competed in back propagation neural networks (BPNN) under the guideline of automatic differentiation mechanism [25] to rapidly reconstruct the multi-order differential formulation. Nevertheless, despite the advantages of evading manufactured prior solutions and exploiting the physical essence of PDEs, the governing equations of copious natural systems can be burdensome to explicitly capture. In this sense, data-driven regime is necessarily implemented to generalize parametric PDEs whose prior computational conditions and solutions are composed into scientific data. To address the aforementioned issues and generalize the data-driven regime beyond the training dataset, operator learning was proposed [26].

Operator learning addresses the challenges introduced by the aforementioned data-driven regime through the application of neural networks, referred to as neural operator [27]. In dealing with parametric PDEs, neural operators learn the complex mapping between the initial condition space and the solution space, both of which are defined as infinite-dimensional functional space. Operator learning obeys the theorem of the universal approximation to nonlinear operators [28]. For a given discrete computational domain, the neural operators infer the pointwise solutions through learning implicit mapping from functional space of parametric initial conditions to solution space. The first neural operator architecture was proposed as DeepONet [26, 29] which composes two neural networks that map the initial condition space and coordinate space separately. One neural network noted as the branch-net is responsible for the function of initial conditions, while the other network noted as the trunk-net is responsible for the coordinates of computational domain. The output of DeepONet is the inner product of the branch-net and trunk-net. DeepONet with its enhanced versions [30-32] is applied to exploit the mechanics problems [33] and mathematical issues [34]. Another type of neural operators utilizes certain integral kernel function to fulfill the stepwise nonlinear mapping between function spaces. Based on the theory of graph neural networks [35], the Graph Neural Operator (GNO) [36] was proposed by learning the integral function with message passing graphs. The Fourier Neural Operator (FNO) [37] transforms the time-dependent input function to Fourier space with Fast Fourier Transform [38]. Inspired by FNO, the Wavelet Neural Operator (WNO) [39] and Multi-wavelet transform-based operator (MWT) [40] utilize the advantage of the wavelets in time-frequency localization of the input space. Specifically, the FNO, WNO and MWT construct kernel integral function through the principle of spectral transform. The aforementioned neural operator architectures were tested on fair data [29] of numerical mechanics problems.

However, two major challenges exist in neural operator architectures for both DeepONet series and spectral transform-based series. (1) State-of-the-art operator architectures [26, 27, 29-32] are constrained by constant spatial discretization schemes and lack the generalization to extrapolate to changeable discretization schemes in computational domains. Once the neural operators are trained by one finite difference scheme, the prediction for new input functions is enforced to provide solutions located at fixed discrete points. This indicates a deficiency in generalizing with diverse discretization schemes that contain massive new discrete points. (2) The spectral transform-based neural operators [37, 39, 40] are mainly efficiently computing in uniformly discretized domain. The uniformly distributed Euclidean space is suitable for discontinuing spectral decomposition [38, 41], yet arbitrary discretization schemes with nonuniform density of discrete points are ignorant. In scenarios where varying levels of precision are required in local computational domain, neural operators fail to capture corresponding solutions. To address the above shortcomings, we propose an enhanced Gaussian kernel-based [42] and graph-based neural operator, referred to as the Dynamic Gaussian Graph Operator (DGGO). Parameterized as neural networks, DGGO constructs state vectors on each hidden layer iteratively into dynamic graph [43] as the integral term of Dynamic Gaussian Graph (DGG) integral kernel. The observation input vectors are composed of arbitrary discrete points and corresponding time-dependent terms defined in general Euclidean space. DGGO comprises three stages, wherein the input space is transformed to the solution space through integral kernel operators. First, the input vectors defined in general Euclidean space are transformed by the forward DGG kernel to high-dimensional uniform metric space, referred to as manifold space. Metric vectors defined in uniform metric space are both frequency and spatial localized to be mapped through spectral transformation. Second, the metric vectors defined in the uniform metric space are mapped to the frequency domain through replaceable spectral transform-based kernel function which is parameterized by Fast Fourier Transform in this work. Finally, metric vectors are mapped back to the general Euclidean space obtaining pointwise solutions through the inverse DGG kernel. Salient contributions concomitant with DGGO are as follows:

(1) For given infinite-dimensional supervised input-output pairs of arbitrary discretization schemes, DGGO is flexible in identifying coordinates of discrete points and stochastic time-dependent functional terms as observation. It outputs pointwise solutions of new unseen discretization schemes of mechanics problems.

(2) DGGO preserves the consistency of spectral transform-based operators on both spatial and frequency domains, offering extra regular constraints on solution space. It extends the compatibility of neural operators to arbitrary non-uniform spatial discretization schemes on mechanics problems with FAIR data.

(3) The principle of DGGO can be implemented to casually combine with any spectral transform-based kernel such as the Fourier kernel. It can be applied as interface for spectral transform-based operators to generalize on arbitrary discrete mechanics problems.

The proposed DGGO is tested across time-dependent or time-independent mechanics problems governed by non-linear PDEs as numerical experiments. Ablation experiments are executed to validate the impact of spatial transformation effect of the DGG kernel. The proposed method is utilized for engineering application as well.

The rest of the paper is organized as follows: the general concept and the standard problem formulation of operator learning are discussed in **Sect. 2**. The methodology of the proposed framework DGGO including the Dynamic Gaussian Graph kernel and the kernel-based operator is introduced in **Sect. 3**. Mechanics problems are used for numerical cases to test the efficacy and robustness of DGGO in **Sect. 4**. Ablation experiments are implemented to study the effectiveness of spatial transformation by DGG kernel in **Sect. 5**. DGGO is utilized to forecast stress field of geometrically variable hyper-elastic material in **Sect. 6**. We discuss overall conclusions in **Sect. 7**.

# 2 Target issues and neural operators

## 2.1 Problem formulation

Neural operator is utilized for mapping between observation vectors defined in infinite-dimensional space and oriented vectors defined in solution space. For a certain computational domain $D$, $\partial D \in \mathbb{R}^d$ governed by a PDE to be solved, the neural operator is responsible for mapping from the functional initial condition vectors $F(u(x,0))$ and functional boundary condition vectors $H(u(\partial D,t))$ defined in input function space to the output $u(x,t)$ defined in solution space. The supervised neural operator learning for solving PDEs is learning the implicit mapping between two infinite dimensional spaces from a finite collection of observed input-output pairs. The initial condition and the boundary condition are observed as prior knowledge to infer the unknown solutions. We define two complete normed vector spaces $\mathcal{A}(D;\mathbb{R}^{d_a})$ and $\mathcal{U}(D;\mathbb{R}^{d_u})$ in Banach space to denote the input and output of the function, respectively. To fulfill the transformation mapping, the non-linear operator $\mathcal{T}^\dagger : \mathcal{A} \to \mathcal{U}$ is parameterized as neural networks to be trained so that the solution vector space of the PDEs $\mathcal{U}(D;\mathbb{R}^{d_u})$ can be attained with different input vector spaces $\mathcal{A}(D;\mathbb{R}^{d_a})$. Specifically, let the input vector $a \in \mathcal{A}$ and solution vector $u \in \mathcal{U}$ form the supervised input-output pairs $\{a_i, u_i\}_{i=1}^N$, where $N$ denotes the number of pairs for any given computational domain. In the PDE that governs a certain computational domain, the degree of discretization determines the number of supervised input-output pairs. The mapping operator $\mathcal{T}^\dagger$ is constructed as

$$\mathcal{T}^\dagger : \mathcal{A} \times \Theta_{NN} \mapsto \mathcal{U}, \tag{1}$$

where $\Theta_{NN}$ is the finite-dimensional parameter space and the $\mathcal{T}(\cdot, \theta_{NN}^\dagger) = \mathcal{T}_{\theta_{NN}^\dagger} \approx \mathcal{T}^\dagger$ is parameterized by $\theta_{NN}^\dagger \in \Theta_{NN}$. For the updating of network parameters $\theta_{NN}^\dagger$, normally the cost function $\mathcal{C} = \mathcal{U} \times \mathcal{U}$ should be minimized as an optimization problem

$$\min_{\theta_{NN} \in \Theta_{NN}} \mathbb{E}\left[\mathcal{C}\left(\mathcal{T}(a, \theta_{NN}), \mathcal{T}^\dagger(a)\right)\right], \tag{2}$$

where $\mathbb{E}(\cdot)$ denotes the maximum likelihood estimation of minimizing the cost function $\mathcal{C}(\mathcal{T}(a, \theta_{NN}), \mathcal{T}(a))$ to determine $\theta_{NN}^\dagger$ in the test-train setting by the data-driven observations input-output pairs $\{a_i, u_i\}_{i=1}^N$. The mapping $u_j = \mathcal{T}^\dagger(a_j)$ is realized by finite-dimensional approximating the operator via neural networks.

It should be noted that by conforming to the solving logistics of PDEs, the input of the mapping operator is the constructed vectors that consist of initial condition, boundary condition and the spatial coordinates of discrete points for a given computational domain. By the mapping of the well-trained operator $\mathcal{T}^\dagger$, we obtain the output as the solution based on the corresponding input condition. The operator is considered to map within the real-valued functions on $\mathbb{R}^d$.

## 2.2 Operator learning via neural networks

The operator $\mathcal{T}^\dagger$ that maps from the input of initial, boundary conditions and coordinates to the output of solutions for the PDE is parametrized by neural networks. Normally, the coordinates condition are extracted from the spatial discretization scheme. The operator learning is implemented by training a deep neural network with the supervised dataset $\{a_i \in \mathbb{R}^{n_D \times d_a}, u_i \in \mathbb{R}^{n_D \times d_u}\}_{i=1}^{N}$ to exploit the non-linear map $\mathcal{T}^\dagger$. The solutions of the discrete computational domain are accessed through point-wise calculation. To evaluate the spatial features, common operator learning let $D_i = \{x_1, \cdots, x_n\} \subset D$ be discretized by $n$ points that are uniformly distributed in the domain $D$, which denotes the resolution of discretization. For each discrete point, the observation vectors $a \in \mathbb{R}^{n_D \times d_a}$ consist of point-wise values of initial condition or boundary condition, combined with the $d$-dimensional coordinates. The neural operator generates the solution $u(x)$ with $x \in D$.

Generally, the neural operator via neural networks is formulated as iterative architecture $v_1 \mapsto v_2 \mapsto \cdots v_j \cdots \mapsto v_l$, where $v_j$ for $j = 1, \cdots, l$ takes values in $\mathbb{R}^{d_v}$. The input $a(x,t) \in \mathcal{A}$ is first up-sampled by a transformation $U(\cdot)$ and obtained as high-dimensional vectors $v_0(x,t) = U(a) \in \mathbb{R}^{d_v}$. In the neural network architecture, the transformation $U(\cdot)$ is designed as a multiple-layer perceptron (MLP) with shallow layers. Each layer in the networks conducts non-linear mapping to update the vectors as $v_{j+1} = L(v_j)$, where the layer-wise function $L(\cdot): \mathbb{R}^{d_v} \mapsto \mathbb{R}^d$ takes values in $\mathbb{R}^{d_v}$. The updates of $v_j$ are performed specifically as

$$v_{j+1}(x) = g\big((K(a;\phi) \cdot v_j(x)) + W \cdot v_j(x)\big); \quad x \in D, j \in [1,l], \tag{3}$$

where $g(\cdot)$ denotes a replaceable non-linear activation function and $W: \mathbb{R}^{d_v} \to \mathbb{R}^{d_v}$ denotes the linear bias set for each layer, respectively. The integral operator $K: \mathcal{A} \times \theta_k \to \mathcal{U}$ is parameterized by $\phi \in \theta_k$ and updated by network training. The kernel operator mapping is defined by

$$\big(K(a;\phi)v_j\big)(x) := \int_D k_\phi(x, y, a(x), a(y); \phi) v_j(y) dy, \quad \forall x \in D, \tag{4}$$

where $k_\phi(\cdot): \mathbb{R}^{2(d+d_a)} \to \mathbb{R}^{d_v \times d_v}$ works as kernel function for each iterative layer, and it is parameterized by the neural networks that are learned by the supervised data. The kernel function can be specifically parameterized by manually designed kernel. An example is parameterizing $k_\phi$ in spectral space using Fast Fourier Transform (FFT). The operator kernel can be replaced by Fourier integral kernel operator as

$$\big(K(a;\phi)v_j\big)(x) = \mathcal{F}^{-1}\big(\mathcal{F}(\kappa_\phi) \cdot (\mathcal{F}v_j)\big)(x), \quad \forall x \in D, \tag{5}$$

where the mapping of neural network $\kappa_\phi$ is parameterized in Fourier space, $\mathcal{F}$ and $\mathcal{F}^{-1}$ denote the forward and inverse Fourier transform. The loss function is leveraged for the best parameters $\theta_{NN}^\dagger$ as

$$\theta_{NN}^\dagger = \arg\min_{\theta_{NN}} \mathcal{L}\big(\mathcal{T}(a, \theta_{NN}), \mathcal{T}^\dagger(a)\big), \tag{6}$$

# 3 Dynamic Gaussian Graph Operator

In this section, we introduce the Dynamic Gaussian Graph Operator (DGGO). DGGO is a generalized kernel-based neural operator that consists of three blocks: (1) forward DGG kernel, (2) spectral transform-based operator, (3) inverse DGG kernel. The input vectors are mapped to the metric vectors defined in the high-dimensional uniform metric space through the forward DGG kernel. Then, the metric vectors are flexible with spectral transform-based operators and are projected to the frequency domain. Finally, the inverse DGG kernel constructs the solution space by mapping the metric vectors to the solution vectors.

## 3.1 Dynamic Gaussian Graph kernel

Arbitrary discrete difference schemes imply multiple non-uniform spatial-temporal discretization of the computational domain. The DGG kernel is responsible for mapping between the generalized Euclidean space and high-dimensional uniform metric space forward and inversely. Vectors defined in the uniform metric space are regarded as latently uniform and can be consistently mapped through spectral transformation to the frequency domain. The framework of the forward DGG kernel connected with the inverse DGG kernel is shown in **Fig. 1**. We define a $n$-point arbitrary discretization scheme $D_r = \{x_1, x_2, \cdots, x_n\} \subseteq D$ in the $d$-dimensional generalized Euclidean space of the original domain $D$. Specifically, each observation discrete point $x \in \mathbb{R}^d$ is defined in Euclidean space with $D, \partial D \in \mathbb{R}^d$ being the $d$-dimensional computational domain and boundary, respectively. Besides, the timing series $t \in \mathbb{R}^{d_t}$ forms the corresponding time-dependent component of the observation. The discrete point $x$ and corresponding time-dependent term $t(x)$ are stacked to form the input vectors $a(x,t(x))\big|_{D_r} \in \mathbb{R}^{n \times d_a}$. In the DGG kernel, the input vectors defined in generalized Euclidean space undergo spatial mapping in the form of finite local neighbor sphere $S_\rho^d$ with each discrete point $x$ as patch center. For a discrete domain $D_r \subseteq D$, we denote input vector space as $\mathcal{A}(D; \mathbb{R}^{d_a})$ defined in Euclidean space and complete normed vector space as $\mathcal{M}(D; \mathbb{R}^{d_\mathcal{M}})$ defined in the uniform metric space which obeys the Banach space properties. To realize the spatial and frequency localized for features of all dimensions on the uniform metric space, the DGG kernel $\mathcal{K}: \mathbb{R}^d \times \mathbb{R}^d \to \mathbb{R}$ aims to realize shift-invariant property (satisfy $\mathcal{K}(x, x') = \mathcal{K}(x - x')$ when referring to as $\mathbb{R}^d \to \mathbb{R}$) and preserve the similarity within local neighbor sphere $S_\rho^d$ on the metric space originating from the input vectors space. Since the input vectors are in the form of point-cloud, we make use of the transform block proposed in the PointNet [44] to align the input vectors to a canonical space by applying a $d_a \times d_a$ matrix. The matrix is constructed through a transformation mapping network $T(\cdot)$. The DGG kernel is formulated as multiple iterative hidden layers in a sequence: $h_1 \mapsto \cdots h_j \cdots \mapsto h_l$, where $h_j$ for $j = 1, \cdots, l$ takes values in $\mathbb{R}^{d_h}$, and represents hidden state spaces between the generalized Euclidean space and high-dimensional uniform metric space. The mapping of forward DGG kernel network $NN_{FDGG}$ that realizes $\mathcal{A}(\cdot) \to \mathcal{M}(\cdot)$ can be mathematically written as

$$m(x') := \mathcal{K}_l \circ \cdots \circ \mathcal{K}_j \circ \cdots \circ \mathcal{K}_1 \circ \{a(x,t(x)) \mid a \in \mathcal{A}(\cdot)\}, \tag{7}$$

where, $m(x')\big|_{D_r} \in \mathbb{R}^{n \times d_\mathcal{M}}$ denotes the $M$-dimensional metric vectors defined in uniform metric space that is composed of shift-invariant metric discrete points $x' = (x'_1, \cdots, x'_\mathcal{M}) \in \mathcal{M}(D; \mathbb{R}^{d_\mathcal{M}})$. The metric discrete point contains $M$-dimensional components that are spatial and frequency localized through spectral transformation. For the kernel integral transformation $\mathcal{K}_j$ between hidden state spaces $h_j \mapsto h_{j+1}$, the kernel integral operator yields

$$h_{j+1}(a_{j+1}) = \int_{a_j}^{a_{j+1}} \mathcal{K}_j(a;\phi)(h_j(a), a, \phi) da, \tag{8}$$

where $h_j(a_j)$ denotes the state vectors $a$ defined in the $j$ th hidden state space $h_j$, $\mathcal{K}_j(a;\phi)$ is a neural network parameterized by $\phi \in \Theta_\mathcal{K}$ that operates non-linear mapping of the state vectors $a \in \mathbb{R}^{n \times d_a}$. After layer-wisely mapping the general Euclidean space to the hidden state space $h_{l-1}$, a continuous integral operation is computed over all hidden state vectors to obtain the last hidden state vector $a_l$ on the $h_l$. Specifically, the continuous integral operator for the last hidden state space is written as

$$h_l(a_l) = \int_{a_{l-2}}^{a_{l-1}} \cdots \int_{a_1}^{a_2} \mathcal{K}(a;\phi)(h(a), a, \phi) da. \tag{9}$$

Therefore, we obtain the $l$ th state vector $a_l$ on the last hidden state space. All hidden state vectors $a_j$, where $j = 1, \cdots, l$ are stacked to construct stacked hidden state vector $a^S = (a_1, \cdots, a_l) \in \mathbb{R}^{l \times d_h}$. Once all the layer-wise integrals are performed, we stack $a_l$ with $a^S$, and project it to the target dimension by a down-sample neural network $\mathcal{D}(\cdot)$ to get metric vector $m(x') = \mathcal{D}((a_l, a^S)) \in \mathcal{M}$. After transforming the initial spatial-temporal vector to uniform metric space $\mathcal{A}(\cdot) \to \mathcal{M}(\cdot)$, we use the DGG kernel inversely to reconstruct the oriented vectors defined in the Euclidean space as $a^*(x, t(x))$. The reconstructed vectors $a^*(x, t(x))$ maintain the same dimension with $a$ in the same space. The inverse process of the DGG kernel that generates vectors defined in initial space is written as

$$\{a^*(x, t(x)) \mid a^* \in \mathcal{A}(\cdot)\} := \mathcal{K}_l^* \circ \cdots \circ \mathcal{K}_{l-1}^* \circ \cdots \circ \mathcal{K}_1^* \circ m(x' \mid \mathcal{M}, \|\cdot\|_\mathcal{M}), \tag{10}$$

where the $\mathcal{K}^*$ denotes the inverse iterative kernel integral transformation in inverse DGG kernel networks $NN_{IDGG}$, $(\mathcal{M}, \|\cdot\|_\mathcal{M})$ denotes the normed uniform metric space. The computational principles of the $NN_{IDGG}$ including the layer-wise kernel integral operator, continuous integral operation are in line with the forward DGG kernel networks $NN_{FDGG}$. It should be pointed out that we simplify the depiction of inverse DGG kernel in **Fig. 1** due to its symmetry with the forward DGG kernel. We leverage the Riemann sum approximating to discretize and efficiently calculate the integral operator and use Gaussian graph kernel operator to describe the interrelation within integral term from state vectors, which are detaily introduced in the following subsections.

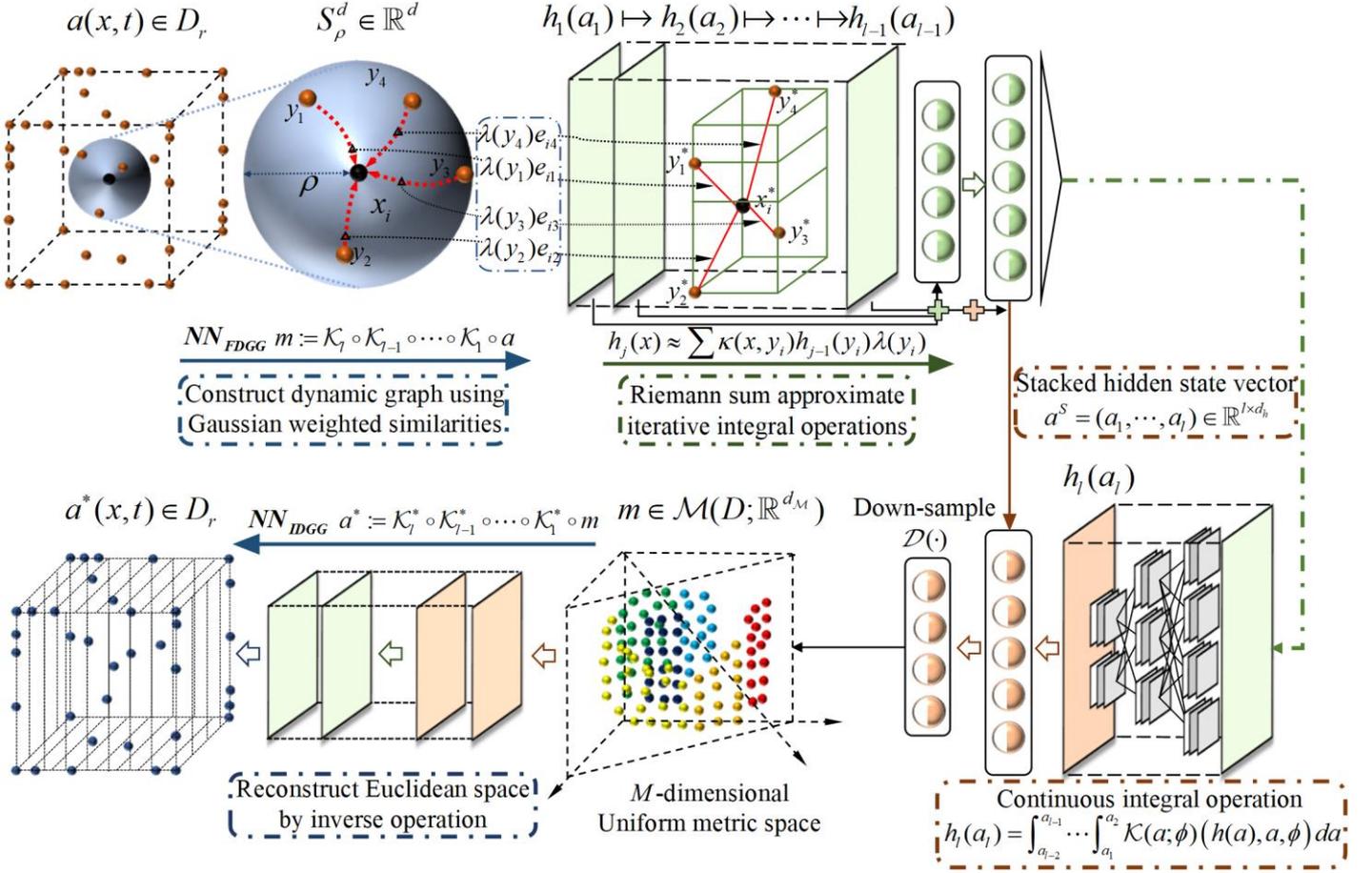

**Fig. 1** Illustration of the architecture of forward and inverse DGG kernel networks

### 3.1.1 Riemann sum approximating of integral operator

In the DGG kernel, the layer-wise mapping can be mathematically represented as integral over the vectors defined in the corresponding hidden state space. By parameterizing the integral operator by neural networks, the updating procedure that represents $h_j \mapsto h_{j+1}$ is written as

$$h_{j+1}(a_{j+1}) = \alpha\left(\omega \cdot h_j(a_j) + \left(\mathcal{K}(a;\phi) \cdot h_j\right)(a_j)\right), \quad \forall a \in D_r, \tag{11}$$

where $\alpha(\cdot): \mathbb{R} \to \mathbb{R}$ and $\omega(\cdot): \mathbb{R}^{d_h} \to \mathbb{R}^{d_h}$ are components of the integral operator that realizes space transform mapping. To fulfill the transformation integral, mapping functions $\alpha(\cdot)$, $\omega(\cdot)$ are non-linear and linear activation functions, respectively. Specifically, the mapping function $\alpha(\cdot)$ is replaceable and the Gaussian Error Linear Units (GELU) function is used to provide smooth and continuous non-linear mapping. The mapping function $\omega(\cdot)$ is selected as $\omega(\xi) = \xi$. Based on the Eqs. (11), the general form of DGG kernel is defined as

$$\left(\mathcal{K}(a;\phi)h_j\right)(x) := \int_{D_r} \Psi_\theta\left(x, y, h(x), h(y); \theta\right) h_j(y) dy, \quad x, y \in D_r, \tag{12}$$

where $\Psi_\theta$ is a learnable kernel realized by dynamic message passing graph to depict the similarity of discrete points on corresponding hidden state space. For the DGG kernel $\mathcal{K}(a;\phi)$ utilized to compute the integral operator in Eqs. (12), neighbor sphere $S_\rho^d \in \mathbb{R}^d$ are truncated within radius of $\rho$ as local domain for integration. The layer-wise update process with $S_\rho^d$ as the integral domain is expressed as

$$h_{j+1}(x) = \int_{S_\rho^d} \Psi_\theta(x, y) h_j(y) dy, \tag{13}$$

where the integral neighborhood sphere $S_\rho^d$ intercepts $d$-dimensional discrete $k$-point within radius of $\rho$ is regarded as integral domain. In the Eqs. (13), the linear mapping term is elusive to simplify the iterative updates. To efficiently calculate the integral operation, we discrete the integral domain by Riemann sum of the sphere on each layer to approximate the integral operations. For each discrete point $x \in \mathbb{R}^d$, the approximate Riemann sum is computed by the nearest $k$-point $y \in \mathbb{R}^d$ on the hidden state space obtained at each layer within radius of $\rho$. The approximate integral computation can be mathematically represented as

$$h_{j+1}(x) \approx \alpha \sum_{i=1}^{k} \Psi_\theta(x, y_i) h_j(y_i) \lambda(y_i), \quad i \in [0, k], \tag{14}$$

where $y_i$ is the neighbor discrete points in the sphere $S_\rho^d = \{x; y_1, \cdots y_k\}$ on hidden state space obtained based on the Euclidean distance metric, $\lambda(y_i)$ denotes the Riemannian coefficients for corresponding integral unit. In the DGG kernel, we parameterize the Riemannian coefficients $\lambda(\cdot)$ by Gaussian kernel function as metric for the similarity of local neighbor sphere. Due to the stability of Gaussian kernel from Euclidean space to high-dimensional manifold space, it preserves the similarity iteratively of $S_\rho^d$ from original Euclidean space, and maps the Euclidean distance to finite-dimensional manifold distance. According to Eqs. (9), the continuous integral operator for the last state space yields

$$h_l(x) = \alpha \sum_{j=1}^{l-1} \sum_{i=1}^{k} \Psi_\theta(x_j, y_{j_i}) a_j(y_{j_i}) \exp(-\frac{\|x_j - y_{j_i}\|^2}{2\sigma^2}), \tag{15}$$

where $\sigma$ is the kernel variance that realizes scaling up or down the reconstructed Euclidean distance. We define the Riemann coefficients as the values of the Gaussian kernel function, noted as Gaussian kernel weighted coefficients. The coefficients between discrete points in the local sphere on each hidden state space are calculated by Gaussian kernel function directly. In the DGG kernel, the Gaussian kernel integral operator $\mathcal{K}(a; \phi)$ is parameterized by $\phi \in \Phi(\sigma, k)$, where the kernel hyperparameters $\Phi$ contain two sub-parameters: $k$ measure the scale of the local neighbor sphere $S_\rho^d$, and kernel variance $\sigma$ works as criteria to determine the Riemannian coefficient $\lambda(\cdot)$. The radius $\rho$ that determines the size of sphere and the number of neighbor discrete points are taken for approximation. Based on Heine's theorem, the error caused by approximating the integral with Riemann sum on a unit length computational domain is denoted as $\mathcal{O}(k^{-2})$. Specifically, the radius and average discrete distances $\delta$ should satisfy $\rho \approx k \cdot \delta$. This indicates that for a given computational domain, the radius for Riemann sum of the sphere is necessarily positively related to $\delta$ and offers a guideline for selecting the hyperparameters $k$ for the integral operation.

### 3.1.2 Dynamic message passing graph

To parameterize the learnable kernel $\Psi_\theta$ that depicts the similarity of discrete points, noted in Eqs. (12), a dynamic message passing graph structure is used to fulfill layer-wise updating the neighbor sphere similarities. The kernel $\Psi_\theta$ is a general graph structure that describes the interrelation within integral term from state vectors. For each hidden state space, kernel $\Psi_\theta(\mathcal{V}, \mathcal{E})$ is parameterized by $\theta \in \Theta_\mathcal{G}$ to represent the neighbor sphere graph, where $\mathcal{V} = \{1, \cdots, k\}$ and $\mathcal{E} \subseteq \mathcal{V} \times \mathcal{V}$ are vertices and edges in graph structure, respectively. The dynamic message passing graph is constructed by discrete points within the radius $\rho$ of local neighbor sphere as $S_\rho^d \in \mathbb{R}^d$, which truncates integral domain, noted in Eqs. (13). It should be pointed out that the graph is self-looped, indicating that vertices are connected to itself and features of center discrete point itself are encompassed to the graph. Specifically, edges in graph are defined as $e_{ij} = \mathcal{G}_\Phi(x_i, x_j)$, where $\mathcal{G}_\Phi$ is the composed of non-linear function that operates the Gaussian graph kernel to compute edges features for message passing for Gaussian kernel graph. The message passing within the local sphere graph is operated by channel-wise symmetric aggregation on edges associated with similarities and vertices of discrete points. We adopt the Gaussian graph kernel operator $\mathcal{G}_\Phi : \mathbb{R}^d \times \mathbb{R}^d \to \mathbb{R}^d$ to depict the Gaussian weighted similarity of discrete points as

$$\mathcal{G}_\Phi(x_i, x_j) = \mathcal{G}_\Phi(x_i, x_j - x_i). \tag{16}$$

This explicitly makes the DGG kernel $\mathcal{K}$ endowed with shift-invariant property. Moreover, the Gaussian graph kernel operator combines the global and local neighbor sphere features to the graph, which are captured by centers $x_i$ and spatial metric $x_j - x_i$, respectively. In particular, we implement the operator $\mathcal{G}_\Phi$ to update the edges of graph and notate it as

$$e_{ijm}^* = \mathrm{GeLU}\left(\theta_m \exp(-\frac{\|x_i - x_j\|^2}{2\sigma^2})(x_j - x_i) + \gamma_m x_i\right), \tag{17}$$

where $\Theta_\mathcal{G} = \{\theta_1, \cdots, \theta_M; \gamma_1, \cdots, \gamma_M\}$ represents the graph updating weights for self-loop edge ($\gamma_m$) and outside-loop edges ($\theta_m$). Therefore, we update the vertices in the form of aggregated form that can be realized by shared MLP as

$$x_{im}^* = \max_{j:(i,j)\in\mathcal{E}} e_{ijm}^*, \tag{18}$$

where the output function $x^* = \max_{j:(i,j)\in\mathcal{E}} \mathcal{G}_\Phi(x_i, x_j)$ is obtained by the $m = \{1, 2, \cdots M\}$ dimension feature updating edges using maximize aggregation with $M$ dimensions in total for each graph on corresponding hidden state space. $x^*$ is invariant to the shift of the input $x_j$, since $\max(\cdot)$ is a symmetric aggregation operation. Therefore, in each iteration within the hidden state space, the Gaussian kernel integral operator $\mathcal{G}_\Phi$ maintains permutation invariance, thereby enabling the DGG kernel to preserve shift invariant property globally. On the $l$ th hidden state space, the graph is noted as $\Psi_\theta^{(l)} = (\mathcal{V}^{(l)}, \mathcal{E}^{(l)})$, where edges noted as $\mathcal{E}^{(l)} = (e_{ij_1}^{(l)}, e_{ij_2}^{(l)}, \cdots e_{ij_k}^{(l)})$ are of the form $(x_i, x_{j_{i1}}^{(l)})$, $(x_i, x_{j_{i2}}^{(l)})$, ..., $(x_i, x_{j_{ik}}^{(l)})$ such that $x_{j_{i1}}^{(l)}$, $x_{j_{i2}}^{(l)}$, ..., $x_{j_{ik}}^{(l)}$ are enclosed within the local neighbor sphere $S_\rho^d$. The construction of dynamic graph is illustrated in **Fig. 2**.

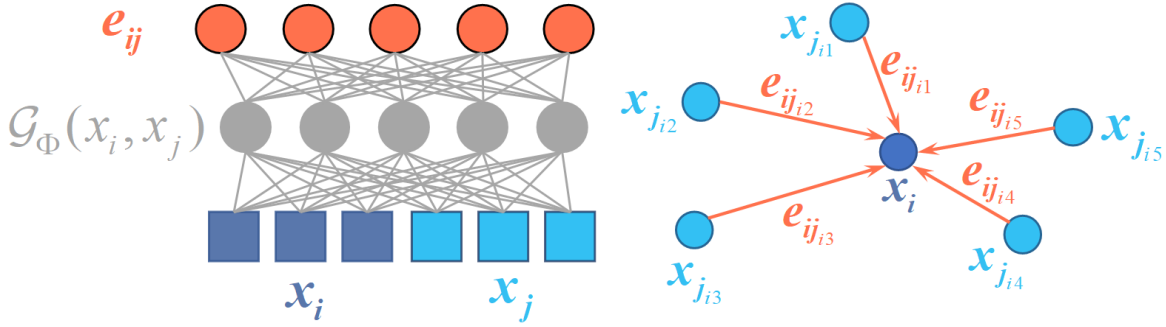

**Fig. 2** Computing edges in graph from a point pair $e_{ij} = \mathcal{G}_\Phi(x_i, x_j)$ and graph aggregation as message passing

### 3.2 DGG kernel-based neural operator

Based on the proposed DGG kernel and operator learning via neural networks, we propose a kernel-based neural operator architecture. We utilize the DGG kernel as the DGG block that works forward and inversely. The forward DGG kernel is responsible for mapping the initial input vectors to high-dimensional uniform metric space, referred to as manifold space. Then, the high-dimensional metric vectors are mapped via spectral transform block which is the neural operator parametrized in Fourier space by using the Fast Fourier Transform. After that, the transformed vectors are reconstructed to the Euclidean space by the inverse DGG kernel. Both input vectors and output vectors are defined in the general Euclidean space. The operator learning process is minimizing the kernel loss function $\mathcal{L}(a^*, a)$. The architecture of DGG kernel-based neural operator (DGGO) is shown in **Fig. 3** and the algorithm of the DGG kernel is presented in **Algorithm 1**.

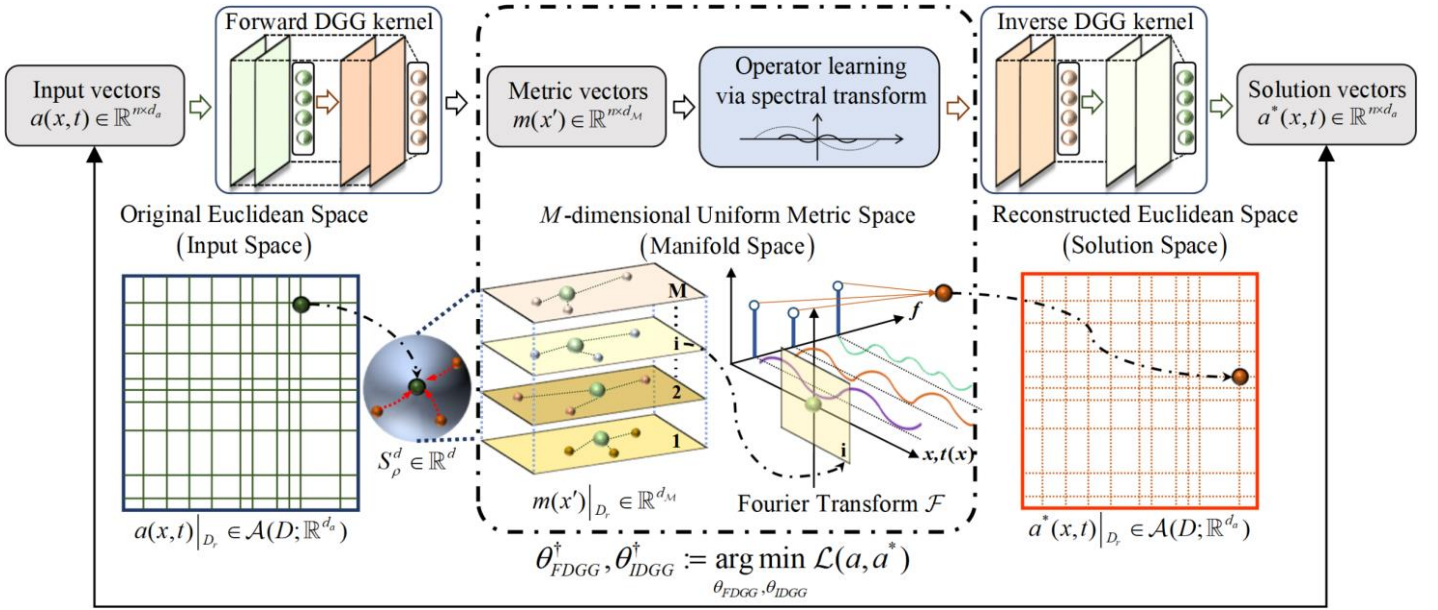

**Fig. 3** Architecture of DGG kernel-based neural operator (DGGO)

**Algorithm 1** Algorithm of the DGG kernel

**Input:** $N$-samples $d$-dimensional arbitrary discrete points $x = (x_1, \cdots, x_n) \in \mathbb{R}^{n \times d}$, time-dependent quantities $t(x) \in \mathbb{R}^{d_t}$, kernel hyperparameter $\Phi = \{\sigma, \rho\}$, graph weight $\Theta = \{\theta_1, \cdots, \theta_M; \gamma_1, \cdots, \gamma_M\}$

**Initialization:** Parameters $\theta_{FDGG}$, $\theta_{IDGG}$ of forward and inverse DGG neural networks $NN_{FDGG}$, $NN_{IDGG}$

1:   Stack quantities to construct observation vectors $a(x, t(x)) \in \mathbb{R}^{d \times d_a}$ on Euclidean space.
2:   **for** epoch $= 1, \cdots, N$ **do**
3:     Construct initial state $a_0(x, t(x))$ and graph Gaussian graph $\Psi_0(x, e \in \mathcal{E})_\Theta$.    ▷ Eqs. (17)
4:     Apply alignment matrix multiply to state $a_0$ via transformation $T(\cdot)$: $a_1 = a_0(x, t(x)) \otimes T(\Psi_0) \in \mathbb{R}^{d \times d_a}$.
5:     **for** $j = 1, \cdots, l$ map through hidden state space iteratively $h_j \mapsto h_{j+1}$ **do**
6:       Obtain the local neighbor sphere $(S_\rho^n)_j = \{x; y_1, \cdots y_k\}$ from the state vector $a_j$ on hidden space $h_j$
7:       Construct hidden state $a_j$ to dynamic graph with Gaussian weighted similarities $\Psi_j(x, e \in \mathcal{E})_\Theta$.
8:       Perform integral operation $h_j(a_j) \mapsto h_{j+1}(a_{j+1}) = \alpha \sum_{i=1}^k (\Psi_j(x, y_i) h_j(y_i) \lambda(y_i))$.    ▷ Eqs. (14)
9:       **if** $j \neq l$ **then**
10:         Compute the dynamic continuous integral: $h_l(a_l) = \alpha \sum_{j=1}^{l-1} \sum_{i=1}^k \Psi_j(S_\rho^n)_j h_j(a_j) \lambda(a_j)$    ▷ Eqs. (15)
11:       **end if**
12:     **end for**
13:     Stack hidden state vectors $a_1 \to a_l$, create aggregated dynamic integral vector $a^S = (a_1, \cdots, a_l)$
14:     Down-sample the $h(a_l)$ using transformation networks $\mathcal{D}(\cdot)$: $m = \mathcal{D}((a_l, a^S)) \in \mathbb{R}^{d_\mathcal{M}}$.
15:     Reconstruct quantities vectors with inverse DGG kernel as final output: $a^*(S^n, t) \in \mathbb{R}^{d \times d_a}$.    ▷ Eqs. (10)
16:     Compute the kernel loss and gradient of the total loss: $\mathcal{L}(a^*, a)$, $\dfrac{\partial \mathcal{L}(a^*, a)}{\partial \theta_{FDGG}} + \dfrac{\partial \mathcal{L}(a^*, a)}{\partial \theta_{IDGG}}$.
17:     Update the parameters $\theta_{FDGG}$, $\theta_{IDGG}$ of the $NN_{FDGG}$, $NN_{IDGG}$ with gradient-based algorithm.
18:   **end for**

**Output:** $M$-dimensional metric vector $m(x')$, where $x' = (x'_1, \cdots, x'_\mathcal{M}) \in \mathcal{M}(D; \mathbb{R}^{d_\mathcal{M}})$ on uniform metric space, parameters $\theta_{FDGG}$, $\theta_{IDGG}$ of forward and inverse DGG neural networks $NN_{FDGG}$, $NN_{IDGG}$.

### 3.2.1 Dynamic Gaussian Graph block

For PDEs to be solved, the observation vector $a(x_1^{In}, \cdots, x_N^{In}) \in \mathbb{R}^{d_a}$ is constructed by a series of arbitrary discrete points extracted from domain $D_r \subseteq D$. Forward DGG kernel is computed as integral operator parameterized by learnable dynamic graph neural networks to obtain metric vectors $m(x^M)$ on uniform metric space. In the Riemann sum approximating of integral operator, the Riemannian coefficient $\lambda(y_i^{In})$ is calculated by the Gaussian kernel function to quantify weighted similarities within the local neighbor sphere $S_\rho^d(x^M) = \{x^{In}; y_1^{In}, \cdots y_k^{In}\}$. For a given $M$-dimensional metric vector $m(x_1^M, \cdots, x_R^M) \in \mathbb{R}^{d_\mathcal{M}}$ defined in uniform metric space, the inverse DGG kernel is computed symmetrically to the forward DGG kernel as Eqs. (10) and thus, we output the vectors $a^*(x_1^{Out}, \cdots, x_N^{Out}) \in \mathbb{R}^{d_a}$ defined in general Euclidean space as solution. The forward and inverse kernel can be regarded as encoder and decoder operations for the kernel-based neural operator, respectively.

The construction of graph and usage of Gaussian kernel function on each layer within the DGG kernel block induces three hyperparameters. The first parameter is the number of discrete nodes $k$ involved in the local sphere that are nearest neighbors to the center points. The nodes of graph are aggregated to the center to identify the computational domain arbitrarily discretized by a given resolution. The second parameter is the Gaussian kernel variance $\sigma$, which directly determines the construction of edges in the graph by computing Riemannian coefficients. The third parameter is the dimension of the high-dimensional metric space. The dimension of the manifold metric space are the features of state vectors mapped from the initial features including time-dependent conditions and coordinates. The specifical hyperparameters of the DGG operator selected against corresponding numerical problems are listed in **Table. 1**.

### 3.2.2 Spectral transform block

The observation input vectors $a(x_1^{In}, \cdots, x_N^{In})$ defined in arbitrary discrete computational domain are transformed to the uniform metric space to form the latently uniformly distributed metric vectors $m(x_1^M, \cdots, x_R^M)$. We implement spectral transform-based operator to the metric vectors before reconstructing the output vectors $a^*(x_1^{Out}, \cdots, x_N^{Out})$ by the inverse DGG kernel operator. The metric vectors $m(x_1^M, \cdots, x_R^M)$ are mapped via iterative architecture $v_1 \mapsto \cdots v_j \cdots \mapsto v_l$, where $v_j$ for $j = 1, \cdots, l$ introduced in **Sect. 2.2**.

In the present work, the spectral transformation integral operator is replaced by a convolution operator defined in Fourier space. Indeed, any wavelet-based spectral transform can be utilized here. The specific definition of the Fourier Neural Operator is discussed in [37]. The generalized wavelet transform is given by

$$\mathcal{S}(\alpha, \beta) = \int_D \Gamma(x) \psi_{\alpha,\beta}(x) dx, \tag{19}$$

where the $\mathcal{S}(\cdot)$ indicates the spectral transform of mapping function $\Gamma(x)$ with scaling and translational parameters $\alpha$ and $\beta$, $\psi_{\alpha,\beta}(x)$ is the orthonormal form of spectral wavelet. Normally, the mapping function $\Gamma(x)$ is parameterized by $\alpha$ and $\beta$ as

$$\psi_{\alpha,\beta}(x) = \frac{1}{\sqrt{a}} \psi\left(\frac{x-\beta}{\alpha}\right), \tag{20}$$

where the spectral transform is decided by waveform function $\psi_{\alpha,\beta}(x)$. Here we select the waveform function $\psi_{\alpha,\beta}(x)$ of its frequency components and ignore the translation parameter. Specifically, $\psi_{\alpha,\beta}(x)$ is replaced by $e^{-2i\pi\langle x,k \rangle}$ to form the Fourier transform. The spectral transform $\mathcal{S}(\cdot)$ that is implemented to metric vectors $m$ can be expressed as

$$\mathcal{S}(\mathcal{F}\Gamma)_j(w) = \int_D \Gamma_j(x) e^{-2i\pi\langle m,w \rangle} dm, \quad \mathcal{S}(\mathcal{F}^{-1}\Gamma)_j(m) = \int_D \Gamma_j(w) e^{2i\pi\langle m,w \rangle} dw, \tag{21}$$

for $j = 1, \cdots, d_a$, where $i = \sqrt{-1}$ denotes the imaginary unit. Considering the shift-invariant property mentioned in **Sect. 3.1**, we choose the operator via spectral transform by Fourier operator as

$$\left(\mathcal{K}(\phi) v_t\right)(m) = \mathcal{F}^{-1}\left(R_\phi \cdot (\mathcal{F} v_t)\right)(m), \quad \forall x \in D, \tag{22}$$

where $R_\phi$ denotes the Fourier transform of a periodic function realized by non-linear neural networks $\kappa_\phi$ which is parameterized by $\phi \in \Theta_\mathcal{K}$ discussed in **Sect. 2.2**. The Fourier spectral transform induces the frequency mode $w \in \mathbb{Z}^d$ as a hyperparameter. On the frequency domain, we assume: $(\mathcal{F}v_t)(k) \in \mathbb{C}^{d_v}$ and $R_\phi(k) \in \mathbb{C}^{d_v \times d_v}$. According to the metric vectors $m(x_1^M, \cdots, x_R^M)$ on the uniform metric space $\mathcal{M}(D; \mathbb{R}^{d_\mathcal{M}})$, the initial arbitrary inequivalent discrete resolution is mapped to $s_1 \times \cdots \times s_d = R$ with $R \in \mathbb{N}$ features on the space. Therefore, the Fast Fourier Transform can be used to efficiently calculate the Fourier transform which is detailly illustrated in Ref. [37]. We refer the reader to this work for further details, since we make use of the FNO just to present one of those combinations of DGG kernel and kernel-based neural operator. Any spectral transformation can be used to replace this block including WNO [39], etc. The related hyperparameter of the spectral transformation block and the dataset size for corresponding numerical problems are illustrated in **Table. 1**.

**Table. 1** Dataset size for numerical problems and structure parameters for DGGO

| Numerical Problems | Datasets | | Mode | Metric dimension | DGGO | | $g(\cdot)$ |
|---|---|---|---|---|---|---|---|
| | Train | Test | | | $\sigma$ | $k$ | |
| 1D Burgers (continuity) | 1000 | 100 | 6 | 32 | 3-10 | 4-14 | GeLU |
| 1D time-dependent wave advection | 1000 | 100 | 6 | 32 | 9-11 | 8-12 | GeLU |
| 2D Darcy flow | 1000 | 100 | 6 | 48 | 4-6 | 3-8 | GeLU |
| 2D time-dependent Navier–Stokes | 1000 | 100 | 6 | 48 | 4-6 | 3-8 | GeLU |

# 4 Operator learning with numerical mechanics problems

To test the efficiency of DGGO, we implement Fourier transform combined with DGG kernel as discussed in **Sect. 3**. Besides, numerical cases of fluid and gas dynamics, phase-field modeling are selected as diverse physical systems. The numerical mechanics problems consist of both one-dimensional and two-dimensional computational domains. The cases we select include hyperbolic, elliptic and parabolic PDEs that involve spatial and time quantities. As comparative experiments, we test the performance of five state-of-the-art neural operators including (1) Fourier Neural Operator (FNO), (2) Wavelet Neural Operator (WNO), (3) DeepONet and (4) POD-DeepONet. Besides, (5) U-Net [45] and (6) PointNet [44], regarded as classic deep neural networks (DNN), are taken as comparison in Case 1-4. Considering the application of DeepONet series methods are restricted within constant discretization schemes, we use them as comparative methods for uniformly discrete mechanical problems [29] in Case 5. The performance of each architecture is measured by point-wise $L^2$ relative error between the prediction and the ground truth. For the optimization algorithm for updating parameters of neural networks, all the testing architectures are trained by AdamW [46] and the initial learning rate is set to be 0.003 with weight decay of $10^{-5}$ every 50 epochs at a rate of 0.75. All numerical cases are trained on NVIDIA GeForce RTX 2080 SUPER 8GB GPU platform. We trained the proposed method and comparative architectures three times on each numerical case and average results are recorded. The relative errors are distributed in the magnitude of $10^{-2}$ at the convergence stage, which is a reasonable level for accuracy. In each training process, we use early-stopping principle to ensure achieving convergence to a controllable extent. We discuss the details of each case and specific comparative results in the following subsections.

## 4.1 Case 1: 1D Burgers equation

Case 1 is the one-dimensional Burgers equation with continuity in the solution field, which is a non-linear PDE modeling the flow of a viscous fluid. It takes the form of periodic boundary condition as

$$\partial_t u(x,t) + \frac{1}{2}\partial_x u^2(x,t) = \upsilon \partial_{xx} u(x,t), \quad x \in (0,1), t \in (0,1]$$

$$u(x=0,t) = u(x=1,t), \quad x \in (0,1), t \in (0,1] \tag{23}$$

$$u(x,0) = u_0(x), \quad x \in (0,1)$$

where $\upsilon \in \mathbb{R}_+$ is the viscosity coefficient, $u_0$ is the initial condition that takes value from $L^2_{\text{per}}((0,1);\mathbb{R})$. In this case, we generate the dataset by letting the initial condition $u_0(x) \sim \mathcal{N}(0, 625(-\Delta + 25I)^{-2})$. As for the periodic condition, we let $u(x-\pi,t) = u(x+\pi,t)$ to satisfy the period of $2\pi$. Setting conditions for Case 1 are taken from Ref. [37], where the viscosity coefficient $\upsilon = 0.1$. The training process is learning the mapping from initial condition $u(x,0)$ to the final state $u(x,1)$, and finalizing DGGO as $\mathcal{K}: u(x,0) \mapsto u(x,1)$. Considering this numerical case uniformly divides the spatial coordinate into 8092 discrete points, we randomly take the resolution of 512, 256, 128, 64 and 48 from the overall discrete points for each input initial condition. Therefore, we have 1100 sets of arbitrary discrete forms including training and testing datasets.

The predicted results by DGGO on all resolutions are illustrated in **Fig. 4**, where five initial conditions $u(x,0)$ and corresponding arbitrary discretization schemes in testing dataset are selected. Based on initial conditions, the predicted solution curves fit in well with the ground truth curves exactly, with seldom outliers. Predicted solutions fit in with the ground truth across all resolutions. Specifically, the comparative results are presented in **Table. 2** for all resolutions. The proposed DGGO has significantly lower error than any other neural operator and DNN across all resolutions.

**Table. 2** Comparative relative error of 1D Burgers equation

| Network Architecture | Spatial resolution | | | | |
|---|---|---|---|---|---|
| | 512 | 256 | 128 | 64 | 48 |
| FNO | 6.56% | 8.49% | 10.27% | 13.37% | 14.83% |
| WNO | 13.17% | 16.62% | 20.36% | 27.94% | 30.89% |
| U-Net | 21.42% | 23.52% | 26.56% | 27.87% | 29.93% |
| PointNet | 6.78% | 8.74% | 9.23% | 12.12% | 13.98% |
| **DGGO** | **2.80%** | **3.71%** | **4.17%** | **4.28%** | **5.22%** |

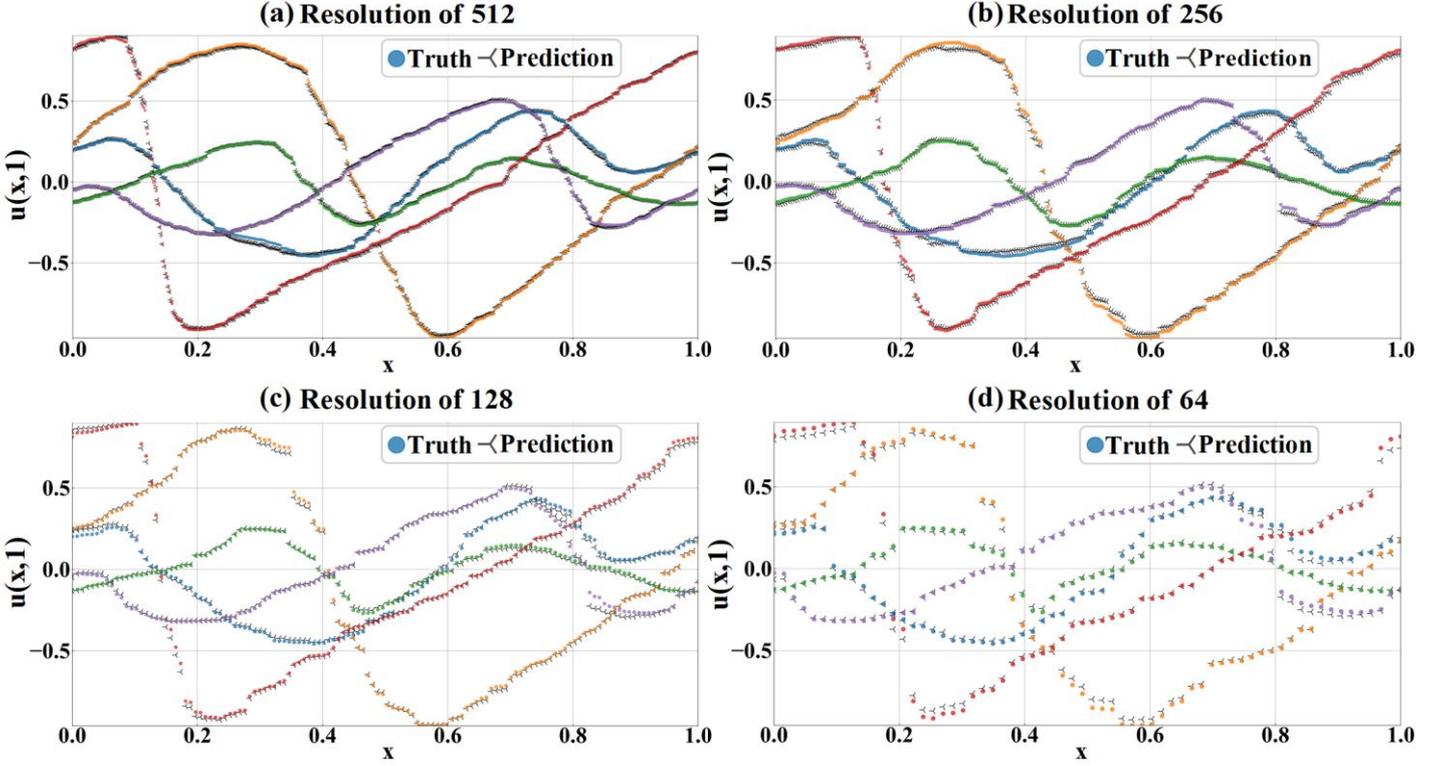

**Fig. 4** Predicted solutions and ground truth of 1D Burgers equation with multiple resolutions

## 4.2 Case 2: 1D time-dependent wave advection equation

Case 2 is a time-dependent hyperbolic PDE that describes the wave advection process in one dimension. We study the oriented solution governed by the wave advection under prior velocity field. The one-dimensional time-dependent wave advection equation with periodic boundary condition yields

$$\partial_t u(x,t) + v \partial_x u(x,t) = 0, \qquad x \in (0,1), t \in (0,1)$$

$$u(x-\pi,t) = u(x+\pi,t), \qquad x \in (0,1), t \in (0,1) \qquad (24)$$

$$u(x,0) = h_{\{c-w/2, c+w/2\}} + \sqrt{\max\left(h^2 - (a(x-c))^2, 0\right)}, \quad x \in (0,1)$$

where $v \in \mathbb{R}_+$ denotes the speed of flow, parameters $\{c, w, h\}$ measure the property of square wave centered at $x = c$ of width $w$ and height $h$ as the initial condition. The specific values of $\{c, w, h\}$ are randomly selected from $[0.3, 0.7] \times [0.3, 0.6] \times [1, 2]$ to form arbitrary square wave condition as input. Considering the spatial-temporal, the field is discretized to 40 points spatially and 40 steps temporally with a time step of 0.025. In this case, we randomly take the resolutions of 36 and 32 from the overall discrete points with each resolution on time-step of 1, 10, 20, and 30 to predict. For $v = 1$, the training process is learning the mapping from initial condition $u(x,0)$ to multiple time-step states and finalizing DGGO $\mathcal{K}: u(x,0) \mapsto u(x,t)$, where $t = 0.025, 0.25, 0.5, 0.75$.

The predicted results by the DGGO on all resolutions are illustrated in **Fig. 5**. We present five discrete solutions $u(x,t)$ derived from corresponding initial condition $u(x,0)$ on multiple time steps under different resolutions. We outline the approximating continuous distribution trend of the solutions to differentiate solutions. It can be noted that the predicted solutions are in line with the ground truth over time. Specifically, the comparative results are presented in **Table. 3** for all resolutions on each time step. The proposed DGGO robustly predicts the solutions in high quality across two resolutions 36 and 32. According to the comparative results, errors are increasing with longer periods from the start. Our DGGO always keeps predicted errors within reasonable interval.

**Table. 3** Comparative relative error of 1D time-dependent wave advection equation

| Network Architecture | Spatial resolution & Time step | | | | | | | |
|---|---|---|---|---|---|---|---|---|
| | 36 | | | | 32 | | | |
| | 1 | 10 | 20 | 30 | 1 | 10 | 20 | 30 |
| FNO | 7.79 % | 9.83 % | 10.89 % | 10.86 % | 12.39 % | 15.83 % | 16.54 % | 15.14 % |
| WNO | 25.13 % | 28.48 % | 28.54 % | 25.30 % | 28.74 % | 34.84 % | 36.92 % | 37.26 % |
| U-Net | 26.74 % | 27.28 % | 28.23 % | 27.98 % | 31.22 % | 40.33 % | 42.85 % | 41.63 % |
| PointNet | 6.20 % | 7.89 % | 8.42 % | 8.37 % | 9.78 % | 10.04 % | 10.36 % | 9.97 % |
| **DGGO** | **3.01 %** | **3.34 %** | **4.07 %** | **4.55 %** | **6.76 %** | **6.69 %** | **7.01 %** | **6.23 %** |

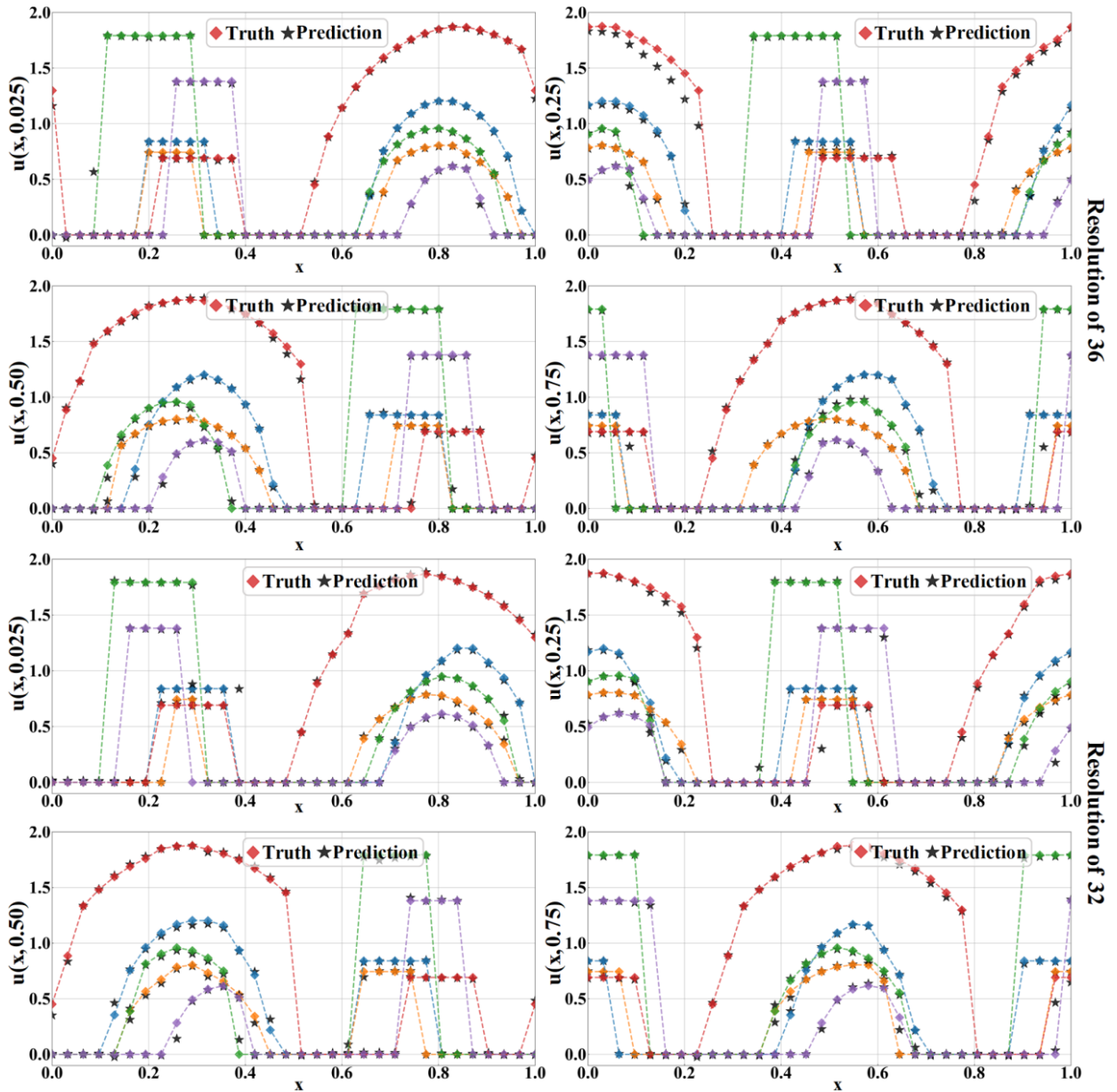

**Fig. 5** Predicted solutions and ground truth of 1D time-dependent wave advection equation

## 4.3 Case 3: 2D Darcy flow equation in a rectangular domain

The proposed method should be implemented to solve more than one-dimensional mechanics problems. Case 3 is the two-dimensional Darcy flow that models the fluid flow through porous media. It can describe both gas and liquid flow in two-dimensional domain by a second-order nonlinear elliptic PDE which forms as

$$\begin{aligned} -\nabla \cdot (a(x,y)\nabla u(x,y)) &= f(x,y), & x,y \in \Omega \\ u(x,y) &= u_0(x,y), & x,y \in \partial\Omega \end{aligned} \quad (25)$$

where $a(x,y)$ is the permeability field, $u(x,y)$ is the pressure field, and $f(x,y)$ is the source function. The source term is selected as constant function with $f(x,y)=1$. The two-dimensional domain is defined in $x \times y \in (0,1)^2$ as a rectangular domain. In this case, $u_0(x,y)=0$ is chosen as the zero Dirichlet boundary condition. Same as in Case 1, we aim to learn the mapping between the preset arbitrary permeability to the pressure field which is $\mathcal{K}: a(x,y) \mapsto u(x,y)$ as a two-dimensional problem. We extract 72, 60, 48, 36, 24 discrete points from the 421 uniformly discrete points to create arbitrary discretization on both two dimensions and the dataset is taken and reconstructed from Ref. [29]. The arbitrary discrete pointes are extracted to create two-dimensional $n \times n$ point-cloud, with $n = 24, 36, 48, 60, 72$.

The predicted pressure field and the ground truth are bicubically interpolated in **Fig. 6**, and pointwise absolute errors are also illustrated to reflect the overall accuracy. Specifically, the comparative results are presented in **Table. 4** for all resolutions. The proposed DGGO maintains the highest accuracy among all comparative methods. Compared to spectral transformed-based methods, the DGGO achieves significant accuracy improvements across all resolutions, indicating that it effectively extends the potency of these methods to non-uniform arbitrary discrete two-dimensional computational domain.

**Table. 4** Comparative relative error of 2D Darcy flow equation in a rectangular domain

| Network Architecture | Spatial resolution | | | | |
|---|---|---|---|---|---|
| | $72 \times 72$ | $60 \times 60$ | $48 \times 48$ | $36 \times 36$ | $24 \times 24$ |
| FNO | 5.15% | 5.89% | 6.35% | 7.13% | 8.50% |
| WNO | 5.78% | 6.34% | 7.12% | 9.28% | 10.88% |
| U-Net | 5.35% | 5.74% | 6.13% | 6.77% | 7.52% |
| PointNet | 4.46% | 4.77% | 5.15% | 5.43% | 5.78% |
| **DGGO** | **3.19%** | **3.42%** | **3.84%** | **4.07%** | **4.48%** |

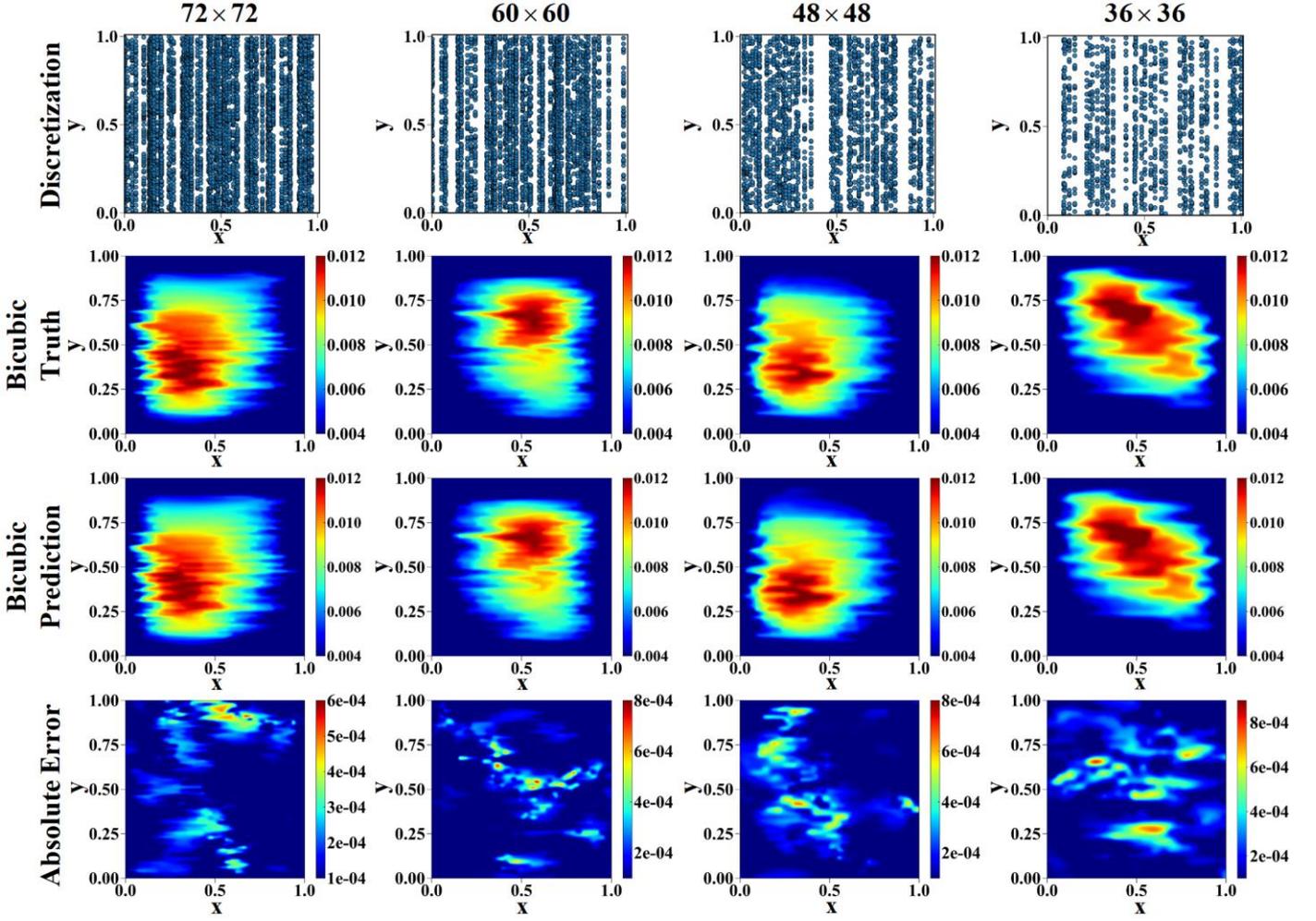

**Fig. 6** Predicted solutions and ground truth of 2D Darcy flow equation from arbitrarily discretized permeability field $a(x,y)$ to $f(x,y)$

## 4.4 Case 4: 2D time-dependent Navier–Stokes equation

We utilize the proposed method to solve the multi-dimensional time-dependent PDE. In Case 4, the Navier-Stokes equation (N-S equation) is used to test in its time-dependent form. The N-S equation is a second-order nonlinear parabolic PDE that describes the property of fluid flow in aerodynamics or thermodynamics. The N-S equation has two versions and one considers the compression effect while the other ignores it. In this case, we consider the two-dimensional incompressible N-S equation in the vorticity-velocity form

$$\begin{aligned}\partial_t \omega(x,y,t) + u(x,y,t) \cdot \nabla \omega(x,y,t) &= \nu \Delta \omega(x,y,t) + f(x,y), & x,y \in (0,1), t \in (0,T] \\ \nabla \cdot u(x,y,t) &= 0, & x,y \in (0,1), t \in (0,T] \\ \omega(x,y,0) &= \omega_0(x,y), & x,y \in (0,1)\end{aligned} \qquad (26)$$

where $f(x,y)$ is the source term, the initial vorticity condition $\omega(x,y,0)$ is denoted as $\omega_0(x,y)$, $\omega(x,y,t)$ and $u(x,y,t)$ are the voracity and velocity, respectively. For the computational domain, $x,y \in [0,1]$. In this case, the viscosity coefficient $\nu \in \mathbb{R}^+$ is chosen as 0.001 and the initial vorticity field $\omega_0(x,y)$ is generated by Gaussian random distribution as $\omega_0 = \mathcal{N}\left(0, 7^{3/2}(-\Delta + 49I)^{-5/2}\right)$. In this case, a time-step of $10^{-4}$ is selected and we aim to learn the map from initial vorticity condition $\omega(x,y,0)$ to multiple time-step vorticity fields.

We aim to finalize DGGO $\mathcal{K}: \omega(t=0) \mapsto \omega(t=t_1,t_2)$, with time-step of 15 and 25. The source term $f(x,y) = 0.1(\sin(2\pi(x+y)) + \cos(2\pi(x+y)))$ is taken as a non-linear function of spatial coordinates. The resolutions are selected to be 48, 40, 32, and 24 on both two dimensions. The original data is taken from Ref. [37], and more processes are finished to arbitrarily acquire discrete points from initial uniform domain. The arbitrary discrete pointes are extracted to create two-dimensional $n \times n$ point-cloud, with $n = 24, 32, 40, 48$.

The bicubically interpolated predicted pressure field and the ground truth are presented in **Fig. 7**, as well as pointwise absolute errors. The quantitative relative error is presented in **Table. 5** for all resolutions. It is noted that the general fluctuation tendency indicates that increase in time step makes prediction more difficult. The DGGO keeps relatively high accuracy even for low resolution and large time-step.

**Table. 5** Comparative relative error of 2D time-dependent Navier–Stokes equation

| Network Architecture | Spatial resolution & Time step | | | | | | | |
|---|---|---|---|---|---|---|---|---|
| | 48×48 | | 40×40 | | 32×32 | | 24×24 | |
| | 15 | 25 | 15 | 25 | 15 | 25 | 15 | 25 |
| FNO | 3.92% | 7.50% | 4.52% | 8.58% | 6.12% | 10.61% | 7.91% | 14.02% |
| WNO | 13.01% | 24.03% | 12.91% | 22.3% | 15.76% | 25.62% | 17.25% | 28.83% |
| U-Net | 8.58% | 12.54% | 7.79% | 13.23% | 8.67% | 14.03% | 10.14% | 16.23% |
| PointNet | 7.63% | 11.28% | 6.32% | 11.61% | 7.98% | 11.47% | 8.79% | 13.71% |
| **DGGO** | **2.33%** | **4.22%** | **2.64%** | **4.86%** | **4.98%** | **6.32%** | **5.84%** | **8.93%** |

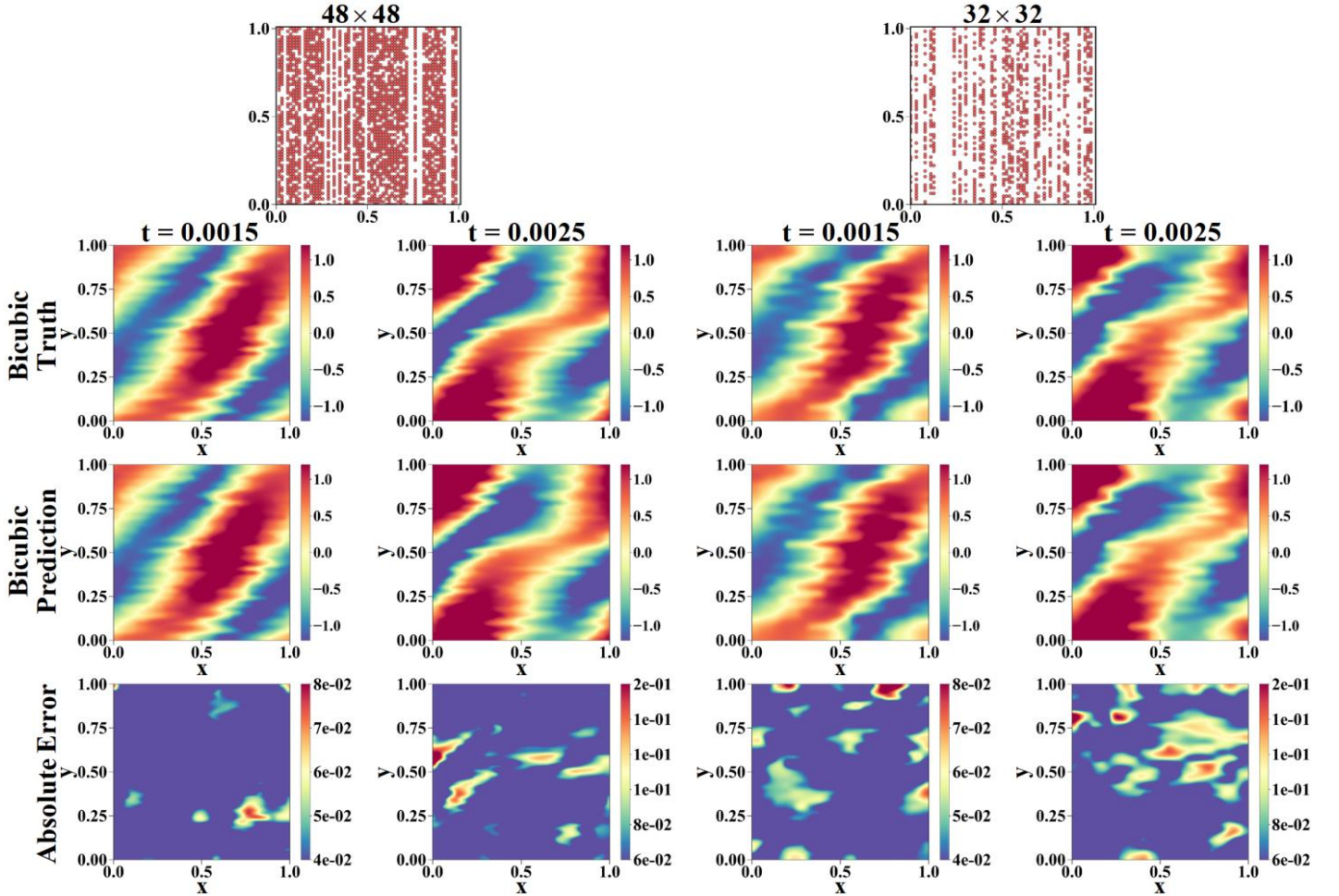

**Fig. 7** Predicted solutions and ground truth of 2D Navier–Stokes equation

## 4.5 Case 5: 1D and 2D uniform discrete schemes

In Case 5, we aim to demonstrate that the DGGO maintains competitive to the state-of-the-art neural operators including the DeepONet series and the spectral-transformed based operators on the classic problems [29, 37]. Uniformly discrete schemes of multiple resolutions are applied for one-dimensional Burgers equation and two-dimensional Darcy flow equation. The resolutions of Burgers equation are selected as 512 and 256, while the resolutions of the Darcy flow equation are selected as $85 \times 85$ and $40 \times 40$. The mean relative errors are represented in **Table. 6**. It is obvious to prove that the DGGO performs reasonably well in the classic uniform discrete schemes for mechanics problems. It proves that our method is suitable for both uniform and non-uniform discretization schemes as a mesh-free method.

**Table. 6** Mean $L^2$ relative error between the truth and predicted results on uniform discrete schemes

| Numerical Problems | Resolution | Network architectures | | | | |
|---|---|---|---|---|---|---|
| | | DeepONet | POD-DeepONet | FNO | WNO | **DGGO** |
| 1D Burgers | 512 | 3.32% | 2.94% | **0.38%** | 2.71% | 0.82% |
| | 256 | 3.21% | 1.96% | **0.45%** | 4.22% | 1.02% |
| 2D Darcy flow | $85 \times 85$ | 3.46% | 2.98% | 1.34% | 1.84% | **1.13%** |
| | $36 \times 36$ | 3.22% | 2.37% | 1.74% | 3.12% | **1.49%** |

## 5 Ablation experiments: Impact of spatial constraint

According to **Algorithm 1**, the training of DGG kernel requires minimizing the kernel loss $\mathcal{L}(a^*, a)$. Specifically, the loss function contains residuals of spatial coordinates $x$ and residuals of time-dependent component $t(x)$ for discrete points before and after mapping through the DGGO. As supervised learning, the constraint on time-dependent term directly guides the DGGO to generate the corresponding target solutions based on given initial conditions. However, the impact of kernel loss on spatial transformation necessitates validation by ablation experiments on with or without constraint on the spatial coordinates. Therefore, we implement ablation experiments on Case 1-4 to detect the impact of spatial constraint. Specifically, loss function of training with spatial constraint contains both residuals of spatial coordinates and time-dependent term, while residual of spatial coordinates is not included in training without spatial constraint. The relative loss of spatial constraint during the learning process is illustrated in **Fig. 8**. It indicates that spatial constraint component of the kernel loss is trained to convergence. The comparative results are shown in **Table. 7**, wherein each case is tested with multiple resolutions or time-step. Training the DGGO with both spatial and time-dependent term yields higher accuracy across all problems. This significantly validates that spatial constraint effectively contribute to spatial transformation and enhance predictive accuracy in the solution space.

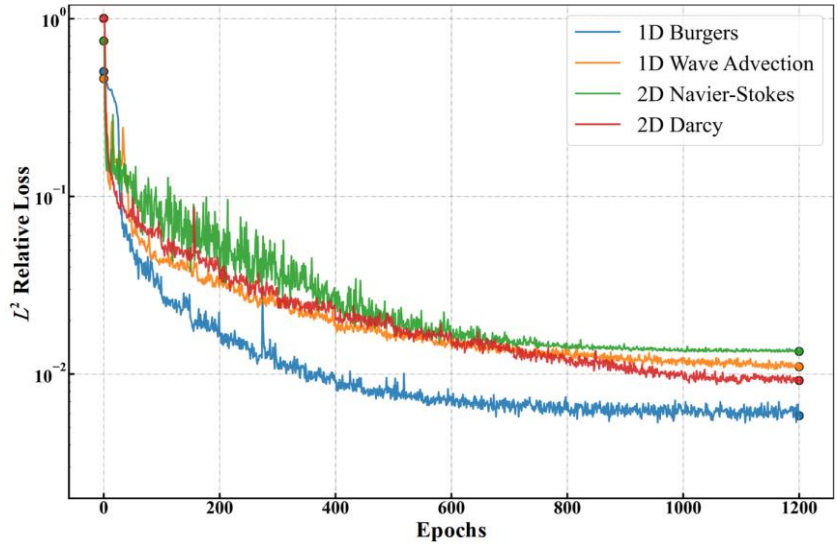

**Fig. 8** Relative loss of spatial constraint

**Table. 7** Comparative accuracy of constraint type (with or without spatial constraint)

| Numerical Problems | Resolution (time step) | Constraint type | |
| --- | --- | --- | --- |
| | | With spatial constraint | Without spatial constraint |
| 1D Burgers | 512 | **2.80%** | 4.18% |
| 1D time-dependent wave advection | 32 (10) | **6.69%** | 8.74% |
| 2D Darcy flow | 48×48 | **3.84%** | 5.71% |
| 2D time-dependent N–S equation | 32×32 (15) | **4.98%** | 6.12% |

  Furthermore, we study the actual effect of the DGG kernel in the spatial transformation and visualize the uniform metric space. The discrete points defined in the uniform metric space are obtained by extracting metric vectors from the output of forward DGG kernel networks. To assess the pointwise spatial distribution, we compare the Euclidean distance matrix of discrete points on the original arbitrary discrete domain and $M$-dimensional vectors defined in uniform metric space of Darcy flow in **Fig. 9**, and Burgers equation in **Fig. 10**. Meanwhile, the distance matrix of uniform discrete points of the domain is given as reference to demonstrate the effectiveness of the DGG kernel on spatial transformation. In **Fig. 9**, at the locations demarcated by the wireframes, a general inference can be concluded that the distance matrix of arbitrary discrete points dissatisfy symmetry on counter diagonal, while uniform discrete points satisfy the property. Through the mapping of the forward DGG kernel, the distance matrix of uniform metric domain satisfies symmetry on counter diagonal, approximating the distribution characteristics of a uniform domain. In **Fig. 10**, we mark the bandwidth on the diagonal where the pair-point distance is close to zero, noted as $l_1$, $l_2$, and $l_3$ on arbitrary discrete domain, uniform metric domain and uniform discrete domain, respectively. Lower bandwidth indicates more uniformly distributed domain, and the relation of $l_1 > l_2 > l_3$ demonstrates that forward DGG kernel effectively transforms the non-uniform discretization space to a latent uniform space that is initially non-uniform discretized.

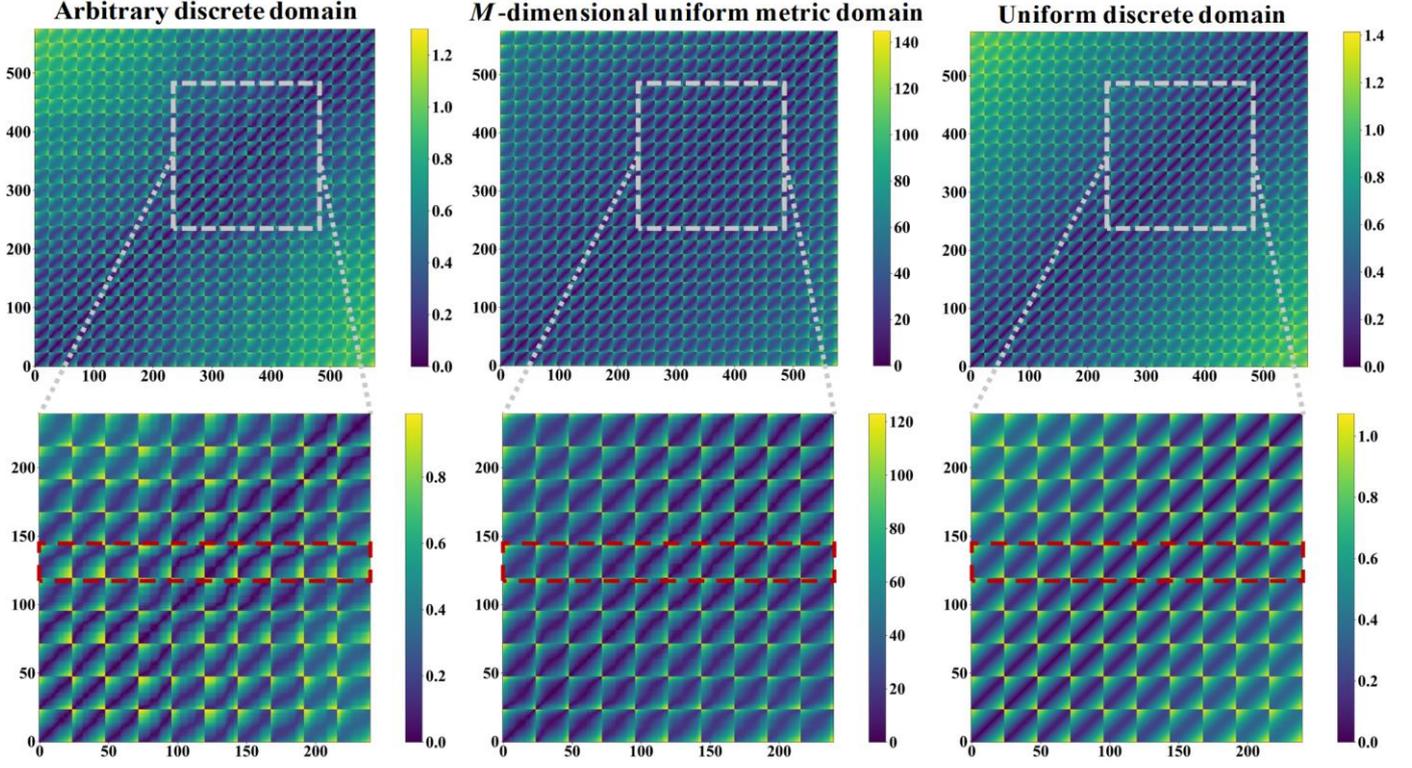

**Fig. 9** Distance matrix of arbitrary discrete domain, $M$-dimensional uniform metric domain, uniform discrete domain, and corresponding local enlargement in 2D Darcy flow (resolution of $24 \times 24$)

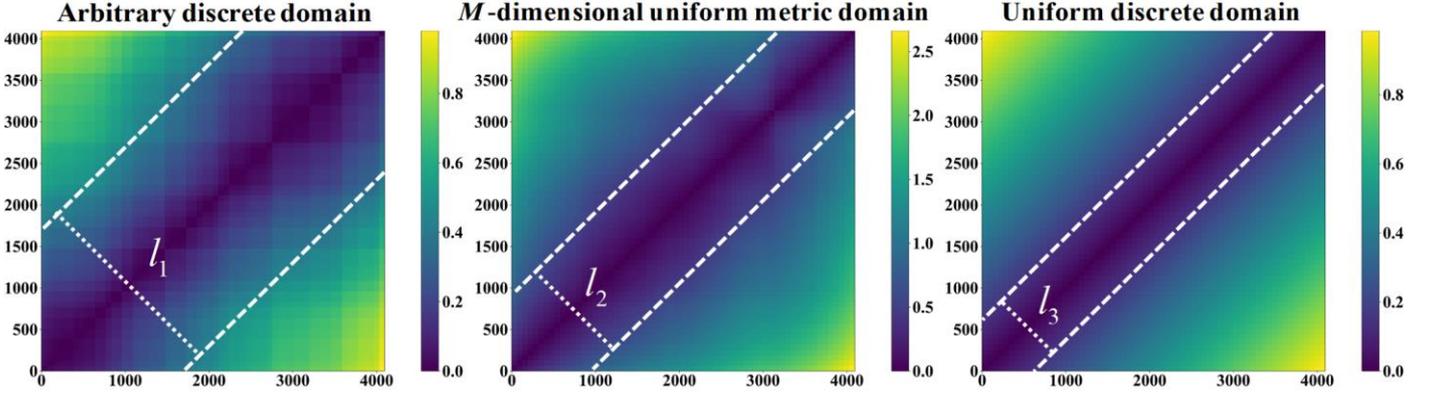

**Fig. 10** Distance matrix of arbitrary discrete domain, $M$-dimensional uniform metric domain, uniform discrete domain, and corresponding local enlargement in 1D Burgers equation (resolution of 64)

## 6 Engineering application: Hyper-elastic material

To furthermore utilize our proposed DGGO on complex structural mechanics application, we thereby consider forecasting the stress of hyper-elastic material [47] in arbitrary discrete computational domain. The governing equation of the solid body yields

$$\rho \frac{\partial^2 \boldsymbol{u}}{\partial t^2} + \nabla \cdot \boldsymbol{\sigma} = 0 \tag{27}$$

where $\rho$ denotes the mass density, $\boldsymbol{u}$ denotes the displacement vector, and $\boldsymbol{\sigma}$ is the stress tensor. In this engineering application, we discuss changeable geometric formations of the solid structure with a constant unit cell domain $D_r = [0,1] \times [0,1]$. Specifically, the changeable geometric formation is realized by constructing arbitrary void of different shapes, shown in **Fig. 11**. The void located at the center of the domain is measured by void radius $r = 0.2 + 0.2/(1+\exp(\tilde{r}))$, where $\tilde{r} \sim \mathbb{N}(0, 4^2(-\nabla + 3^2)^{-1})$. Noticing that the computational domain is simultaneously arbitrarily discretized by mesh-free point-cloud data as the assimilation of discrete points. The unit cell is loaded with tension traction $t = [0, 100]$ on the top edge. The material is selected as incompressible Rivlin-Saunders material, which can be described by the hyper-elastic constitutive model as

$$\sigma = \frac{\partial \omega(\varepsilon)}{\partial \varepsilon} \tag{28}$$
$$\omega(\varepsilon) = C_1(I_1 - 3) + C_2(I_2 - 3)$$

where $I_1 = tr(C)$ and $I_2 = 0.5[tr(C)^2 - tr(C^2)]$ are scalar invariants of the right Cauchy Green stretch tensor $C = 2\varepsilon + 1$ and energy density function parameters $C_1 = 1.86 \times 10^5$, $C_2 = 9.79 \times 10^3$. The task for the DGGO to address is using the discrete points and corresponding void radius as observation input vectors to forecast the pointwise stress field of the unit cell. The training dataset and testing dataset are in size of 2000 and 100, respectively, by implementing finite element solver to obtain the ground truth.

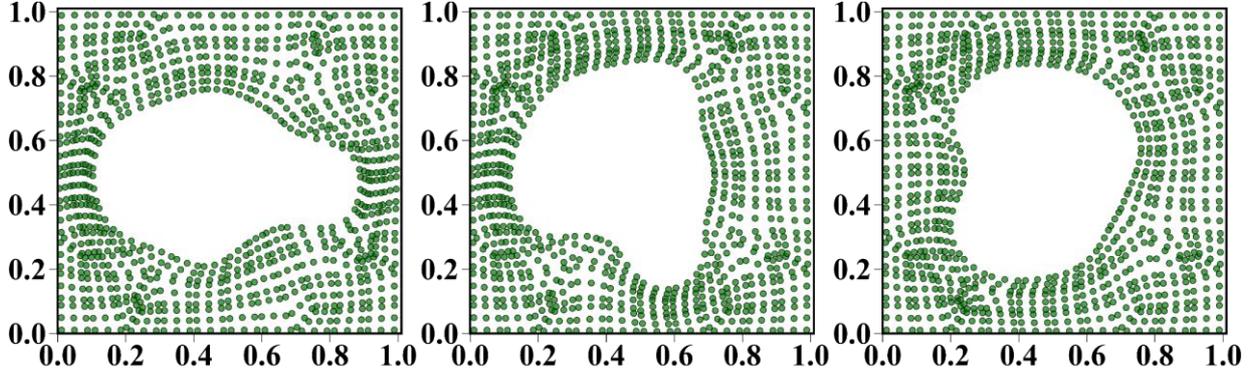

**Fig. 11** Arbitrary discrete computational domain with geometrically variable void located at center

The methodology of forecasting stress of geometrically variable hyper-elastic material is illustrated in **Fig. 12**. In the forecasting task of the pointwise stress field, the DGGO is trained on the training dataset, where 2000 void shapes and corresponding discrete points are included. The input vectors are composed of coordinates of 972 points and identifying sequence of length 42 that differentiates the void shape, while the solutions vectors are the pointwise stress field of the unit domain. We utilize the well-trained DGGO to forecast testing dataset with 100 unseen shapes of the void as validation. The $L^2$ relative error robustly maintains around 1.52%. General prediction results are presented in **Fig. 13**, compared with the ground truth. Once trained, the DGGO can identify diverse spatial discretization and geometric features, generating reasonably accurate stress field.

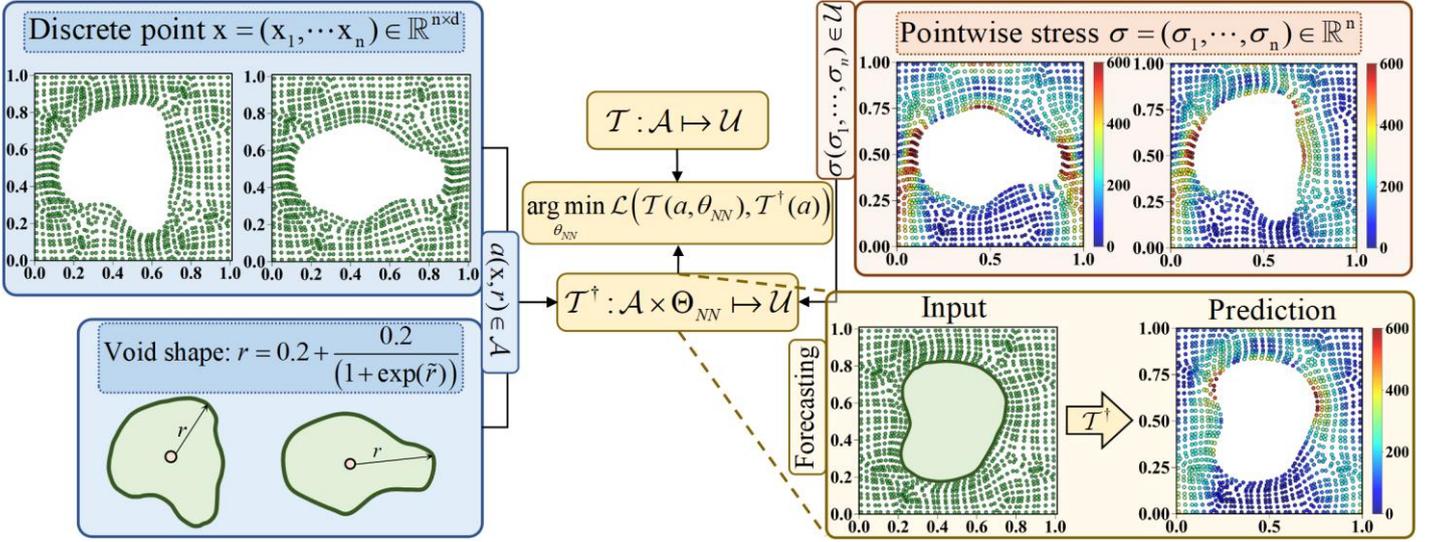

**Fig. 12** Framework of training and testing to forecast the stress field of unit cell by DGGO

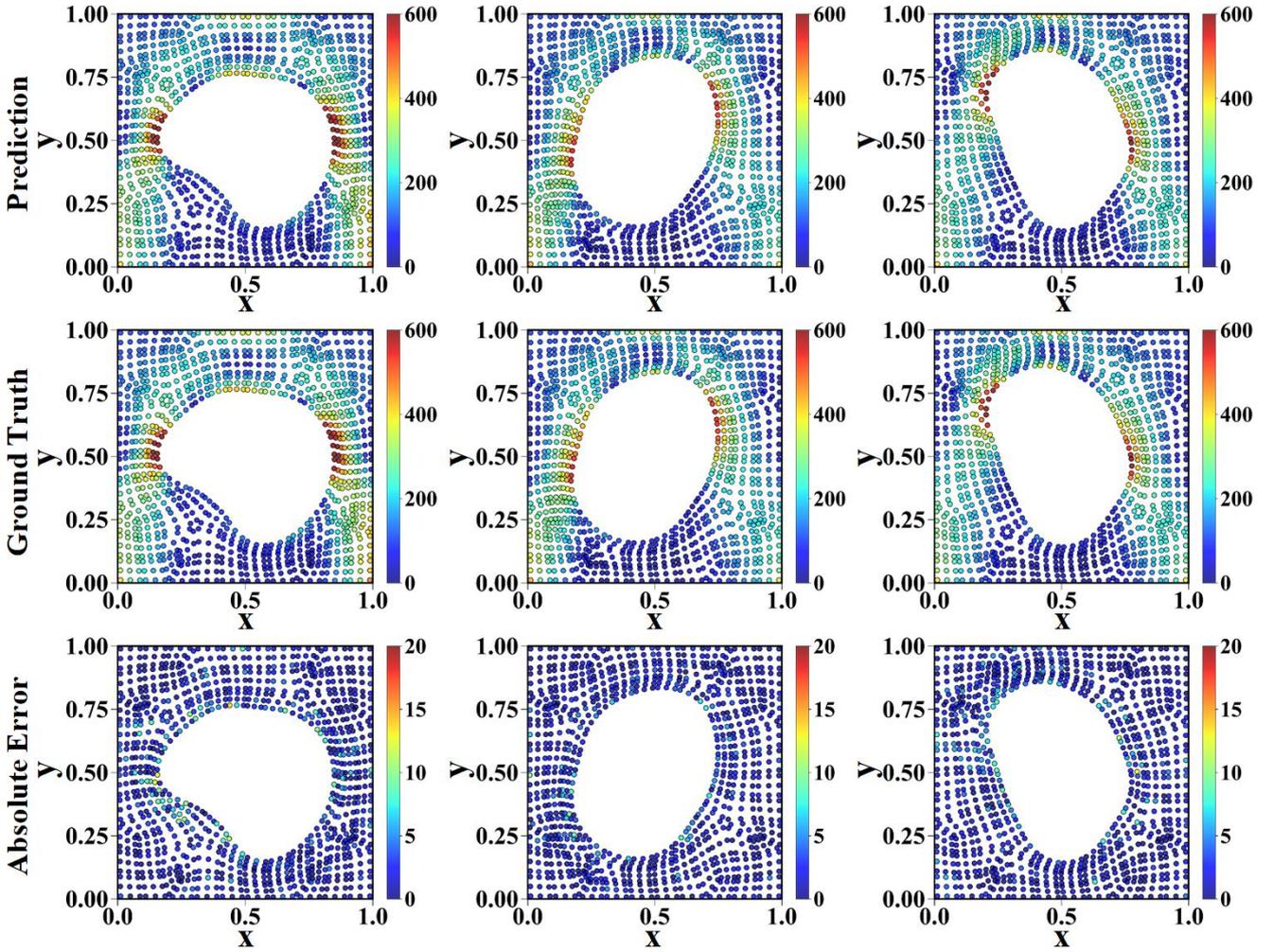

**Fig. 13** Prediction and ground truth results with corresponding absolute error
25

# 7 Conclusions

This article proposed an enhanced neural operator for learning integral kernel operators for non-linear parametric partial differential equations in arbitrary discrete mechanics problems and engineering application. Mathematically, DGGO performs integral kernel on the input observation vectors that contain non-uniform discrete points and corresponding time-dependent term. The proposed DGGO consists of DGG kernel operator that works forward and inversely and replaceable spectral transformed-based operator, which is selected as FNO in this work. The input vectors defined in the general Euclidean space are mapped to the metric vectors defined in high-dimensional uniform metric space through the forward DGG kernel. The DGG kernel operator is iteratively formulated as Gaussian kernel weighted Riemann sum approximating the integral operator and using the message passing graph to depict the interrelation within the integral term on each hidden state space. The DGG kernel operator is parameterized by graph-based neural networks with nonlinear activation functions to fulfill spatial transformation. After being mapped to the uniform metric space, the vectors are regarded as defined in latent uniform domain and transformed by FNO to the spectral domain. The utilization of spectral transform-based operator ensures that metric vectors are localized on both spatial and frequency domains, thus offering external constraints on solution space for parametric PDEs. Finally, the metric vectors are projected to solution vectors through the inverse DGG kernel.

In terms of application, the DGGO is applied to solve both one-dimensional and two-dimensional hyperbolic, elliptic and parabolic PDEs that describe mechanics systems. The comparative state-of-the-art methods are selected to include FNO, WNO, DeepONet, POD-DeepONet, U-Net and PointNet. General results indicate that in comparison with the spectral transform-based operators, the DGGO significantly promotes the predicting accuracy on numerical cases, thereby expanding the application of these methods in learning parametric PDE in arbitrary discrete mechanics problems. Besides, the DGGO maintains competitive on numerical problems with classic uniformly discrete schemes. The quantitative results show that relative errors are at a reasonable magnitude of $10^{-2}$ across all resolutions and time-step on every numerical case, outperforming other operator learning approaches. Ablation experiments are implemented to validate the impact of the spatial constraint in kernel loss of the DGGO. The quantitative results suggest that the kernel loss including residuals of spatial coordinates and time-dependent component enhance the accuracy of operator learning. Furthermore, extraction of the distance matrix of arbitrary discrete domain and uniform metric domain from the forward DGG kernel proves the effectiveness of spatial transformation for shift invariant property and uniformity.

The DGGO demonstrates mighty superiority in engineering application for forecasting the stress field of hyper-elastic material with geometrically variable void located in center. The well-trained DGGO predicts the pointwise stress field of the target unit cell with new unseen shapes of void accurately and the relative error achieves $1.52\%$. The aforementioned cases reflect the flexibility of DGGO in numerical and engineering mechanics problems.

## Declaration of Competing Interest

The authors declare that they have no known competing financial interests or personal relationships that could have appeared to influence the work reported in this paper.